\documentclass{article}

\usepackage[final]{Styles/neurips_2024}

\usepackage[utf8]{inputenc} 
\usepackage[T1]{fontenc}    
\usepackage{hyperref}       
\usepackage{url}            
\usepackage{booktabs}       
\usepackage{amsfonts}       
\usepackage{nicefrac}       
\usepackage{microtype}      
\usepackage{xcolor}         

\usepackage{comment}
\usepackage{graphicx}
\usepackage{duckuments}
\usepackage{bm}
\usepackage{amsmath}
\usepackage[capitalize,noabbrev]{cleveref}
\usepackage{bbm}
\usepackage{enumitem}
\setlist[itemize]{leftmargin=0.2in}
\usepackage{algorithm,algorithmic}
\usepackage{wrapfig}
\usepackage{tcolorbox}
\usepackage{subcaption}
\usepackage{tabularx}
\usepackage{threeparttable} 
\usepackage{multirow}

\DeclareSymbolFont{extraup}{U}{zavm}{m}{n}
\DeclareMathSymbol{\varheart}{\mathalpha}{extraup}{86}
\DeclareMathSymbol{\vardiamond}{\mathalpha}{extraup}{87}

\title{Kaleidoscope: Learnable Masks for Heterogeneous Multi-agent Reinforcement Learning}

\author{
  Xinran Li \;\;\;\;\;\;  Ling Pan \;\;\;\;\;\; Jun Zhang \\ \\
  Department of Electronic and Computer Engineering \\
  The Hong Kong University of Science and Technology \\
  \texttt{
    xinran.li@connect.ust.hk, 
    lingpan@ust.hk, 
    eejzhang@ust.hk
  }
}

\begin{document}

\maketitle

\begin{abstract}
  In multi-agent reinforcement learning (MARL), parameter sharing is commonly employed to enhance sample efficiency. However, the popular approach of full parameter sharing often leads to homogeneous policies among agents, potentially limiting the performance benefits that could be derived from policy diversity. To address this critical limitation, we introduce \emph{Kaleidoscope}, a novel adaptive partial parameter sharing scheme that fosters policy heterogeneity while still maintaining high sample efficiency. Specifically, Kaleidoscope maintains one set of common parameters alongside multiple sets of distinct, learnable masks for different agents, dictating the sharing of parameters. It promotes diversity among policy networks by encouraging discrepancy among these masks, without sacrificing the efficiencies of parameter sharing. 
  This design allows Kaleidoscope to dynamically balance high sample efficiency with a broad policy representational capacity, effectively bridging the gap between full parameter sharing and non-parameter sharing across various environments. 
  We further extend Kaleidoscope to critic ensembles in the context of actor-critic algorithms, which could help improve value estimations.
  Our empirical evaluations across extensive environments, including multi-agent particle environment, multi-agent MuJoCo and StarCraft multi-agent challenge v2, demonstrate the superior performance of Kaleidoscope compared with existing parameter sharing approaches, showcasing its potential for performance enhancement in MARL. The code is publicly available at \url{https://github.com/LXXXXR/Kaleidoscope}.\looseness=-1
\end{abstract}


\section{Introduction}

Cooperative multi-agent reinforcement learning (MARL) has demonstrated remarkable effectiveness in solving complex real-world decision-making problems across various domains, such as 
resource allocation~\citep{marl_dis}, package delivery~\citep{marl_delivery}, autonomous driving~\citep{MARL_autonomous_driving}, and robot control~\citep{marl_robot_swamy2020scaled}. To mitigate the challenges posed by the non-stationary and partially observable environments typical of MARL~\citep{MARL_23_survey}, the centralized training with decentralized execution (CTDE) paradigm~\citep{CTDE} has become prevalent, inspiring many influential MARL algorithms such as MADDPG~\citep{MADDPG}, COMA~\citep{COMA}, MATD3~\citep{MATD3}, QMIX~\citep{QMIX}, and MAPPO~\citep{MAPPO}. \looseness=-1

\begin{wrapfigure}{r}{0.4\textwidth} \vspace{-.2in}
\centering
\includegraphics[width= \linewidth]{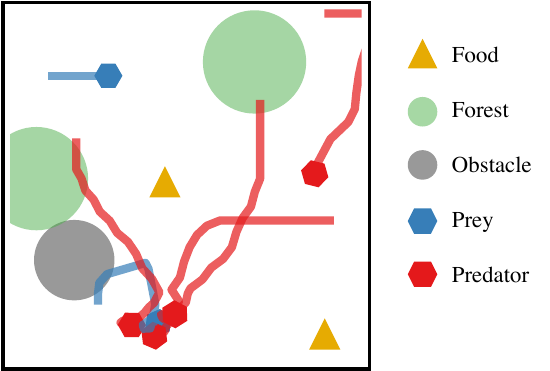}
\vspace{-.2in}
\caption{Full parameter sharing confines the policies to be homogeneous. In this example, all predators pursue the same prey, neglecting another prey in the game \texttt{World}. Further game details are in \cref{supp: env_details}. \looseness=-1}
\label{fig: intro_example}
\vspace{-10pt}
\end{wrapfigure}

Under the CTDE paradigm, parameter sharing among agents is a commonly adopted practice to improve sample efficiency. However, identical network parameters across agents often lead to homogeneous policies, 
restricting diversity in behaviors and the overall joint policy representational capacity. This limitation can result in undesired outcomes in certain situations~\citep{SePS,revisiting,snp}, as shown in \cref{fig: intro_example}, impeding further performance gains. An alternative approach is the non-parameter sharing scheme, where each agent possesses its own unique parameters. Nevertheless, while this method naturally supports heterogeneous policies, it suffers from reduced sample efficiency, leading to significant training costs. This is particularly problematic given the current trend towards increasingly large model sizes, with some scaling to trillions of parameters~\citep{LLM_survey, gpt4_tech_report}. Therefore, it is imperative to develop a parameter sharing strategy that enjoys both high sample efficiency and broad policy representational capacity, potentially achieving significantly enhanced performance. While several efforts~\citep{SePS, snp} have explored partial parameter sharing initiated at the start of training, such initializations can be challenging to design without detailed knowledge of agent-specific environmental transitions or reward functions~\citep{SePS}.

In this work, we build upon insights from previous studies~\citep{SePS, revisiting, snp} and introduce \emph{Kaleidoscope}, a novel adaptive partial parameter sharing scheme. It maintains a single set of policy parameters and employs multiple learnable masks to designate the shared parameters. Unlike earlier methods that depend on fixed initializations, Kaleidoscope dynamically learns these masks alongside MARL parameters throughout the training process. This end-to-end training approach inherently integrates environmental information, and its adaptive nature enables Kaleidoscope to dynamically adjust the level of parameter sharing based on the demands of the environment and the learning progress of the agents. The learnable masks facilitate a dynamic balance between full parameter sharing and non-parameter sharing, offering a flexible trade-off between sample efficiency and policy representational capacity through enhanced heterogeneity. Initially, we build Kaleidoscope upon agent networks, where it achieves diverse policies. 
Following this success, we extend it to multi-agent actor-critic algorithms to encourage heterogeneity among the central critic ensembles for further performance enhancement. \looseness=-1


\begin{tcolorbox}
Just like a \emph{kaleidoscope} uses the reflective properties of rotating mirrors to transform simple shapes into beautiful patterns, our proposed method leverages learnable masks to map a single set of parameters into diverse policies, thereby enhancing task performance.
\end{tcolorbox}

We summarize our contributions as follows:
\begin{itemize}
    \item To enable policy heterogeneity among agents for better training flexibility, we adapt the soft threshold reparameterization (STR) technique to learn distinct masks for different agent networks while only maintaining one set of common parameters, 
    effectively balancing between full parameter sharing and non-parameter sharing mechanisms.
    \item To enhance policy diversity among agents, we introduce a novel regularization term that encourages the pairwise discrepancy between masks. Additionally, we design resetting mechanisms that recycle masked parameters to preserve the representational capacity of the joint networks.
    \item Through extensive experiments on MARL benchmarks, including multi-agent particle environment (MPE)~\citep{MADDPG}, multi-agent MuJoCo (MAMuJoCo)~\citep{facmac_mamujoco} and StarCraft multi-agent challenge v2 (SMACv2)~\citep{smacv2}, we demonstrate the superior performance of Kaleidoscope over existing parameter sharing approaches. 
\end{itemize}

\section{Background}
\paragraph{Multi-agent reinforcement learning (MARL)}
In MARL, a fully cooperative partially observable multi-agent task is typically formulated as a decentralized partially observable Markov decision process (dec-POMDP)~\citep{pomdp}, represented by a tuple $\mathcal{M} = \langle \mathcal{S}, A, P, R, \Omega, O, N, \gamma \rangle$. Here, $N$ denotes the number of agents, and $\gamma \in (0, 1]$ represents the discount factor. At each timestep $t$, with the environment state as $s^t \in \mathcal{S}$, agent $i$ receives a local observation $ o_i^t \in \Omega$ drawn from the observation function $O(s^t,i)$ and then follows its local policy $\pi_i$ to select an action $a_i^t \in A$. Individual actions form a joint action $\bm{a}^t \in A^N$, leading to a state transition to the next state $s^{t+1} \sim P(s^{t+1}| s^t, \bm{a}^t)$ and inducing a global reward $r^t = R(s^t, \bm{a}^t)$. The overall team objective is to learn the joint policies $\bm{\pi} = \langle \pi_1, \ldots, \pi_N \rangle$ such that the expectation of discounted accumulated reward $G^t = \sum_t \gamma^t r^t$ is maximized. \looseness=-1

To learn such policies $\bm{\pi}_{\bm{\theta}}$, various MARL algorithms~\citep{MADDPG,COMA,QMIX,MAPPO} have been developed. For instance, the off-policy actor-critic algorithm MATD3~\citep{MATD3} serves as an example method. Specifically, the critic networks are updated by minimizing the temporal difference (TD) error loss
\begin{equation}
     \mathcal{L}_c(\phi) = \mathbb{E}_{(s^t, \bm{o}^t,  \bm{a}^t, r^t, s^{t+1}, \bm{o}^{t+1}) \sim \mathcal{D}} \left[ \left( y^t - Q (s^t, \bm{a}^t; \phi) \right)^2 \right],
\end{equation}
with 
\begin{equation}
    y^t = r^t + \gamma \min_{j=1,2} Q(s^{t+1}, \pi_1(o_1^{t+1};\theta_1') + \epsilon, \ldots, \pi_N(o_N^{t+1};\theta_N') + \epsilon; \phi_j),
\end{equation}
where $\phi$ are the parameters for critics, $\theta$ are the parameters for actor policies and $\theta'$ are the parameters for target actor policies. And $\epsilon$ is the clipped Gaussian noise, given as $\text{clip}(\mathcal{N}(0, \sigma), -c, c)$.


The policy is updated by the deterministic policy gradient algorithm~\citep{deterministic_pg} 
\begin{equation}
    \nabla \mathcal{J}(\theta_i) = \mathbb{E}_{(s^t, \bm{o}^t,  \bm{a}^t, r^t, s^{t+1}, \bm{o}^{t+1}) \sim \mathcal{D}} \left[ \nabla_{\theta_i} \pi_i(o_i^t;\theta_i) \nabla_{a_i}Q(s^t, a_1, \ldots, a_N \vert_{a_i = \pi_i(o_i^t;\theta_i)}; \phi_1 ) \right].
    \label{eq: matd3_a_obj}
\end{equation}

\paragraph{Soft threshold reparameterization (STR)} Originally introduced in the context of model sparsification, STR~\citep{STR} is an unstructured pruning method that achieves notable performance without requiring a predetermined sparsity level. Specifically, STR applies a transformation to the original parameters $W$ as follows
\begin{equation}
    \mathcal{S}_g(W, s) = \text{sign}(W) \cdot \text{ReLU}\left(\lvert W \rvert - g(s) \right),
\end{equation}
where $s$ is a learnable parameter, $\alpha = g(s)$ serves as the pruning threshold, and $\text{ReLU}(\cdot) = \max(\cdot, 0)$. 
The original supervised learning problem modeled by 
\begin{equation}
    \min_{\bm{W}} \mathcal{L}(\bm{W}; \mathcal{D})
\end{equation}
with $\mathcal{D}$ as the data is now transferred to 
\begin{equation}
    \min_{\bm{W}, \bm{s}} \mathcal{L}(\mathcal{S}_g(\bm{W}, \bm{s}); \mathcal{D}).
\end{equation}
Overall, this approach optimizes the learnable pruning threshold alongside the model parameters, facilitating dynamic adjustment to the sparsity level during training.

\section{Learnable Masks for Heterogenous MARL} \label{sec: method}

\begin{figure}
  \centering
  \includegraphics[width=\linewidth]{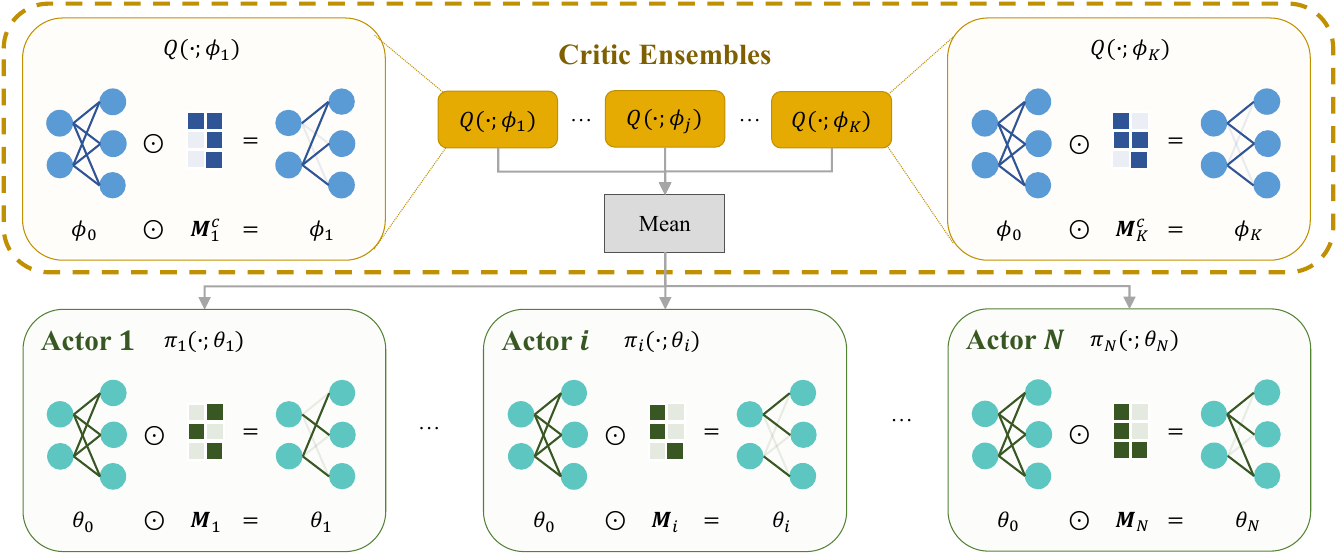}
  \caption{Overall network architecture of Kaleidoscope. It maintains one set of parameters $\theta_0$ with $N$ sets of masks $\left[\bm{M}_i\right]^N_{i=1}$ for actor networks, and one set of parameters $\phi_0$ with $K$ sets of masks $\left[\bm{M}_j^c\right]^K_{j=1}$ for critic ensemble networks, where $N$ is the number of agents, $K$ is the number of ensembles, and $\odot$ denotes the Hadamard product. \looseness=-1}
  \label{fig: arch}
\end{figure}

In this section, we propose using learnable masks as a low-cost method to enable network heterogeneity in MARL. The core concept, illustrated in \cref{fig: arch}, is to learn a single set of shared parameters complemented by multiple masks for distinct agents, specifying which parameters to share.

Specifically, in \cref{sec: str_for_partial_sharing}, we first adapt STR into a dynamic partially parameter sharing method, unlocking the joint policy network’s capability to represent diverse policies among agents. In \cref{sec: policy_div}, we actively foster policy heterogeneity through a novel regularization term based on the masks. Given that the masking technique could excessively sparsify the network, potentially diminishing its representational capacity, in \cref{sec: reset}, we propose a straightforward remedy to periodically reset the parameters based on the outcomes of masking, which additionally mitigates primacy bias. Finally, in \cref{sec: critics}, we explore how to further extend this approach within the critic components of actor-critic algorithms to improve value estimations in MARL and further boost performance. \looseness=-1

For the sake of clarity, we integrate the proposed Kaleidoscope with the MATD3~\citep{MATD3} algorithm to demonstrate the concept within this section. Nevertheless, as a versatile partial parameter-sharing technique, our method can readily be adapted to other MARL algorithms. We defer its integration with other MARL frameworks to \cref{suppl: training_obj_QMIX} and will evaluate them empirically in \cref{sec: experiments}. \looseness=-1

\subsection{ Adaptive partial parameter sharing $\vardiamond$} \label{sec: str_for_partial_sharing}

The core idea of this work is to learn distinct binary masks $\bm{M}_i$ for different agents to facilitate differentiated policies, ultimately aiming to improve MARL performance. To achieve this, we apply the STR~\citep{STR} technique to the policy parameters with different thresholds dedicated to each agent:
\begin{equation}
    \theta_i = \theta_0 \odot \bm{M}_i,
\end{equation}
where $\theta_i$ parameterizes the policy for agent $i$, $\theta_0$ is the set of learnable parameters shared by all agents, and $\bm{M}_i$ is the learnable mask for agent $i$. Specifically, assume $\theta_0 = \left[\theta^{(1)}_0, \ldots, \theta^{(N_a)}_0 \right]$, $\theta_i = \left[\theta^{(1)}_i, \ldots, \theta^{(N_a)}_i\right]$ and $\bm{M}_i = \left[m^{(1)}_i, \ldots, m^{(N_a)}_i\right]$ with $N_a$ being the total parameter count of an agent's network. In line with STR, we compute each element $m_i^{(k)}$ of $\bm{M}_i$ as $m_i^{(k)} = \mathbbm{1} \left[\lvert \theta^{(k)}_0 \rvert > \sigma(s_i^{(k)}) \right]$, where $\sigma(\cdot)$ denotes the Sigmoid function.

The benefits of such a combination are summarized as follows:
\begin{itemize}
    \item \textbf{Preservation of original MARL learning objectives:} Unlike most of the methods in pruning literature, which primarily aim to minimize the discrepancies between pruned and unpruned networks in terms of weights, loss, or activations~\citep{sparsity_survey, efficient_dl_survey, model_compression_survey}, STR maintains the original goal of minimizing task-specific loss, aligning directly with our objectives to enhance MARL performance.
    \item \textbf{Flexibility in sparsity:} Many classical pruning methods require predefined per-layer sparsity levels~\citep{RigL, edge_popup}.
    Such requirements can complicate our design, with the goal not to gain extreme sparsity but rather to promote heterogeneity through masking. The STR technique is ideal in our case as it does not require predefining sparsity levels, allowing for adaptive learning of the masks. \looseness=-1
    \item \textbf{Enhanced network representational capacity:} 
    Utilizing learnable masks for adaptive partial parameter sharing enhances the network’s representational capacity beyond traditional full parameter sharing. In full parameter sharing, agents' joint policies are parameterized as $\bm{\pi}^{\text{ps}}(\cdot \rvert \theta_0) = \langle \pi_1(\cdot \rvert \theta_0), \ldots, \pi_N(\cdot \rvert \theta_0) \rangle$. In contrast, our proposed adaptive partial parameter sharing mechanism parameterizes the joint policies as $\bm{\pi}^{\text{Kaleidoscope}}(\cdot \rvert \theta_0, \bm{M}) = \langle \pi_1(\cdot \rvert \theta_0 \odot \bm{M}_1), \ldots, \pi_n(\cdot \rvert \theta_0 \odot \bm{M}_N) \rangle$. In the extreme case where all the values in $\bm{M}_i$ are $1$s, the function set represented by $\bm{\pi}^{\text{Kaleidoscope}}(\cdot \rvert \theta_0, \bm{M})$ degrades to that of $\bm{\pi}^{\text{ps}}(\cdot \rvert \theta_0)$. In other scenarios, it is a superset of that represented by $\bm{\pi}^{\text{ps}}(\cdot \rvert \theta_0)$. 
\end{itemize}

\subsection{Policy diversity regularization $\clubsuit$} \label{sec: policy_div}

While independently learned masks enable agents to develop distinct policies, without a specific incentive, these policies may still converge to being homogeneous.
To this end, we propose to explicitly encourage agent policy heterogeneity by introducing a diversity regularization term maximizing the weighted pairwise distance between network masks, which is defined as
\begin{equation}
    \mathcal{J}^{\text{div}}(\bm{s}) = \sum_{i = 1, \ldots, n} \sum_{\substack{{j = 1, \ldots, n}\\ j\neq i}}  \lVert  \theta_0 \odot (\bm{M}_i - \bm{M}_j) \rVert_1.
    \label{eq: a_div}
\end{equation}

This term is inherently non-differentiable due to the indicator function $\mathbbm{1}[\cdot]$ inside $\bm{M}$. To overcome this difficulty, following established practices in the literature~\citep{STE,STE_tanh}, we utilize a surrogate function for gradient approximation:
\begin{equation}
    \frac{\partial \mathcal{J}^{\text{div}}}{\partial g(\bm{s}_i)} = - \text{tanh}\left[\frac{\partial \mathcal{J}^{\text{div}}}{\partial \bm{M}_i} \right].
\end{equation}

We formally provide the overall training objective for actors in \cref{suppl: training_obj_MATD3}.

\subsection{Periodically reset $\spadesuit$} \label{sec: reset}
\begin{figure}
  \centering
  \begin{subfigure}[t]{.48\textwidth}
    \centering
    \includegraphics[width=\textwidth]{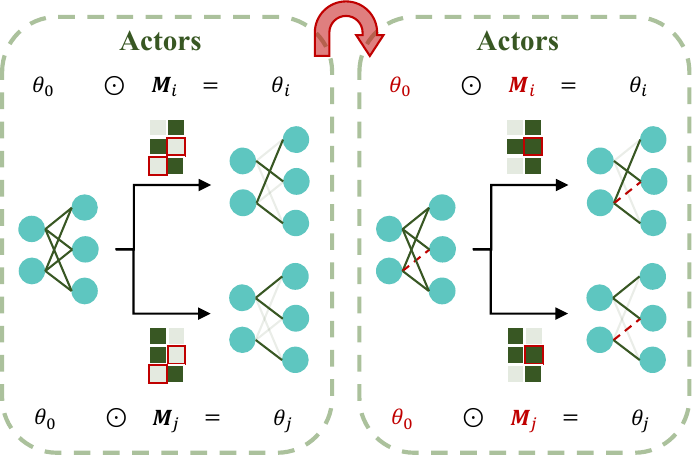}
    \caption{Actors: reinitialize the weights that are masked by all agents with probability $\rho$.}
    \label{fig: reset}
   \end{subfigure}
   \hfill
  \begin{subfigure}[t]{.48\textwidth}
    \centering
    \includegraphics[width=\textwidth]{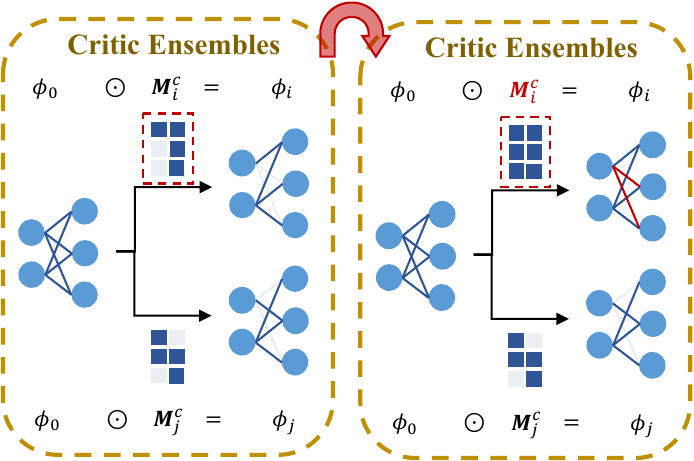}
    \caption{Critic ensembles: reset one set of masks at a time.}
    \label{fig: reset_c}
   \end{subfigure}
  \caption{Illustration on resetting mechanisms.}
  \label{fig: reset_all}
\end{figure}

As the training with masks proceeds, we observe an increasing sparsity in each agent's network, potentially reducing the overall network capacity.
To remedy the issue, we propose a simple approach to periodically reset the parameters that are consistently masked across all $\bm{M}_i$ with a certain probability $\rho$, which is illustrated in \cref{fig: reset}.
At intervals defined by $t \mod \text{reset\_interval} == 0$, if the parameter index $k$ satisfies $\forall i, m^{(k)}_i == 0$, we apply the following resetting rule
\begin{align}
    \theta^{(k)}_0, s^{(k)}_1, \ldots, s^{(k)}_N  \leftarrow 
    \begin{cases}
    \text{Reinitialize}[\theta^{(k)}_0, s^{(k)}_1, \ldots, s^{(k)}_N] &\text{with probability } \rho \\
    \theta^{(k)}_0, s^{(k)}_1, \ldots, s^{(k)}_N &\text{with probability } 1 - \rho
    \end{cases}
    .
\end{align}



This resetting mechanism recycles the weights masked as zeros by all the masks, preventing the networks from becoming overly sparse. A side benefit of this resetting mechanism is the enhancement of neural plasticity~\citep{understanding_plasticity,plasticity_injection}, which helps alleviate the primacy bias~\citep{primacy_bias} in reinforcement learning. Unlike methods that reinitialize entire layers resulting in abrupt performance drops~\citep{primacy_bias}, our resetting approach selectively targets weights as indicated by the learnable masks, thus avoiding significant performance disruptions, as shown in \cref{sec: experiments}. \looseness=-1

\subsection{Critic ensembles with learnable masks} \label{sec: critics}
In actor-critic algorithm frameworks, we further apply Kaleidoscope to central critics as an efficient way to implement ensemble-like critics. By facilitating dynamic partial parameter sharing, Kaleidoscope enables heterogeneity among critic ensembles. Furthermore, by regularizing the diversity among critic functions, we can control ensemble variances. This approach is elaborated in subsequent paragraphs. 

\paragraph{$\vardiamond$ Adaptive partial parameter sharing for critic ensembles} In the standard MATD3 algorithm~\citep{MATD3}, two critics with independent parameters are maintained to mitigate overestimation risks. However, using separate parameters typically results in a low update-to-data (UTD) ratio~\citep{dropout_q}. To address this issue, we propose to enhance the UTD ratio by employing Kaleidoscope parameter sharing among ensembles of critics. Specifically, we maintain a single set of parameters $\phi_0$ and $K$ masks $\left[\bm{M}_j^c\right]^K_{j=1}$ to distinguish the critic functions, resulting in $K$ ensembles $\left[Q(\cdot; \phi_j)\right]_{j=1}^K$ with $\phi_j = \phi_0 \odot \bm{M}_j^c$. 

To be specific, we update the critic networks by minimizing the temporal difference (TD) error loss
\begin{equation}
     \mathcal{L}_c(\phi_j) = \mathbb{E}_{(s^t, \bm{a}^t, s_{t+1}) \sim \mathcal{D}} \left[ \left( y^t - Q (s^t, \bm{a}^t; \phi_j) \right)^2 \right],
     \label{eq: c_j_obj}
\end{equation}
with 
\begin{equation}
    y^t = r^t + \gamma \min_{j=1,\ldots, K} Q(s^{t+1}, \pi_1(o_1^{t+1};\theta_1') + \epsilon, \ldots, \pi_n(o_N^{t+1};\theta_N') + \epsilon; \phi_j).
\end{equation}

And the policies are updated by the mean estimation of the ensembles as 
\begin{equation}
    \nabla \mathcal{J}(\theta_i) = \mathbb{E}_{s^t \sim \mathcal{D}} \left[ \nabla_{\theta_i} \pi_i(o_i^t;\theta_i) \nabla_{a_i} \frac{1}{K} \sum_{j=1}^K \left[  Q(s^t, a_1, \ldots, a_N \lvert_{a_i = \pi_i(o_i^t;\theta_i)}; \phi_j ) \right] \right].
\end{equation}


\paragraph{$\clubsuit$ Critic ensembles diversity regularization} As in \cref{sec: policy_div}, we also apply diversity regularization to critic masks to prevent critics functions from collapsing to identical ones. The diversity regularization to maximize for the critic ensembles is expressed as
\begin{equation}
    \mathcal{J}^{\text{div}}_c(\bm{s}^c) = \sum_{i = 1, \ldots, K} \sum_{\substack{{j = 1, \ldots, K}\\ j\neq i}}  \lVert  \phi_0 \odot (\bm{M}_i^c - \bm{M}_j^c) \rVert_1.
    \label{eq: c_div}
\end{equation}

Intuitively, as training progresses, this term encourages divergence among the critic masks, leading to increased model estimation uncertainty. This process fosters a gradual shift from overestimation to underestimation. As discussed in prior research~\citep{dropout_q,maxmin_q,random_q,ada_q_ensemble}, overestimation can encourage exploration, beneficial in early training stages, whereas underestimation alleviates error accumulation~\citep{TD3}, which is preferred in the late training stage. We formally provide the overall training objective for critic ensembles in \cref{suppl: training_obj_MATD3}.


\paragraph{$\spadesuit$ Periodically reset} To further promote diversity among critic ensembles and counteract the reduction in network capacity caused by masking, we implement a resetting mechanism similar to that described in \cref{sec: reset}. In particular, we sequentially reinitialize the masks $\bm{M}_j^c$ following a cyclic pattern, as illustrated in \cref{fig: reset_c}. In this way, each critic function's mask is trained on distinct data segments, leading to different biases.

In summary, by adopting Kaleidoscope parameter sharing with learnable masks, we establish a cost-effective implementation for critic ensembles that enjoy a high UTD ratio. Through enforcing distinctiveness among the masks, we subtly control the differences among critic functions, thereby improving the value estimations in MARL.


\section{Experimental Results} \label{sec: experiments}
In this section, we integrate Kaleidoscope with the value-based MARL algorithm QMIX and the actor-critic MARL algorithm MATD3, and evaluate them across eleven scenarios in three benchmark tasks. 

\subsection{Experimental Setups}
\paragraph{Environment descriptions} We test our proposed Kaleidoscope on three benchmark tasks: MPE~\citep{MADDPG}, MaMuJoCo~\citep{facmac_mamujoco} and SMACv2~\citep{smacv2}. For the discrete tasks MPE and SMACv2, we integrate Kaleidoscope and baselines with QMIX~\citep{QMIX} and assess the performance. For the continuous task MaMuJoCo, we employ MATD3~\citep{MATD3}. We use five random seeds for MPE and MaMuJoCo and three random seeds for SMACv2, reporting averaged results and displaying the $95\%$ confidence interval with shaded areas. The chosen benchmark tasks reflect a mix of discrete and continuous action spaces and both homogeneous and heterogeneous agent types, detailed further in \cref{supp: env_details}. \looseness=-1

\paragraph{Baselines} In the following, we compare our proposed Kaleidoscope with baselines~\citep{SePS, snp}, as listed in \cref{tab: methods}. For both Kaleidoscope and the baselines, in scenarios with fixed agent types (MPE and MaMuJoCo), we assign one mask per agent. For SMACv2, where agent types vary, we assign one mask per agent type. We use official implementations of the baselines where available; otherwise, we closely follow the descriptions from their respective papers, integrating them into QMIX or MATD3. Hyperparameters and further details are provided in \cref{suppl： suppl_network_arch_hyper}.

\begin{table}[t]
\caption{Methods compared in the experiments. Here, ``adaptive'' indicates whether the sharing scheme evolves during training.}
\label{tab: methods}
\begin{center}
\begin{tabular}{llllp{0.34\linewidth}}
\toprule
Methods             & Paradigm & Sharing level & Adaptive & Descriptions                                                        \\ \midrule
NoPS                & No sharing        & -                       & No              & Agents have distinct parameters                         \\
FuPS                & Full sharing      & Networks                & No              & Agents share all the parameters                           \\
FuPS + ID                & Full sharing      & Networks                & No              & Agents share all the parameters with agent IDs in input                      \\
SePS                & Partial sharing         & Networks                & No              & Agents are clustered to share parameters within each cluster            \\
MultiH                & Partial sharing         & Layers                & No              & Agents share all the parameters except for distinct action heads            \\
SNP                & Partial sharing          & Neurons                 & No              & Agents share specific neurons based on fixed, random pruning \\
Kaleidoscope & Partial sharing          & Weights                 & Yes             & Agents share parameters based on distinct, learnable masks          \\ \bottomrule
\end{tabular}
\end{center}
\end{table}

\subsection{Results}
\paragraph{Performance}

\begin{figure}[t]
\centering
\raisebox{-\height}{\includegraphics[width=0.95\textwidth]{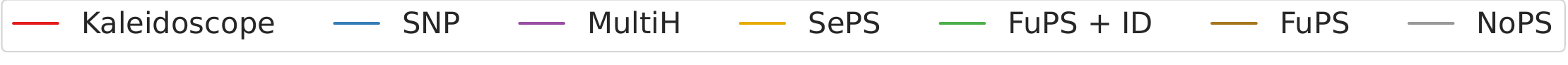}}
\par
\begin{subfigure}[t]{.24\textwidth}
    \centering
    \includegraphics[width=\textwidth]{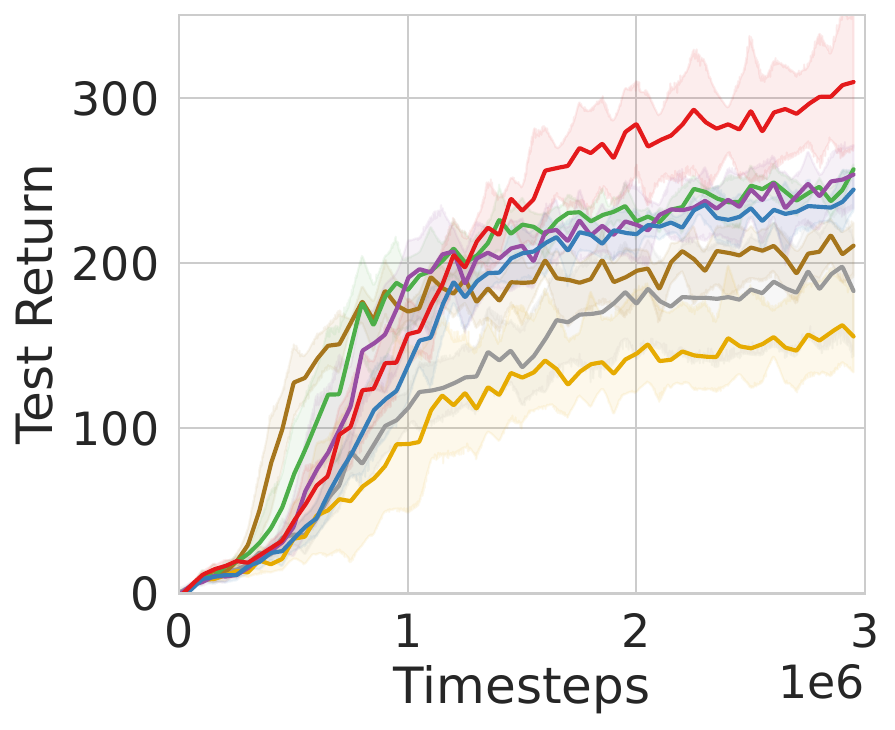}
    \caption{World}
\end{subfigure}
\begin{subfigure}[t]{.24\textwidth}
    \centering
    \includegraphics[width=\textwidth]{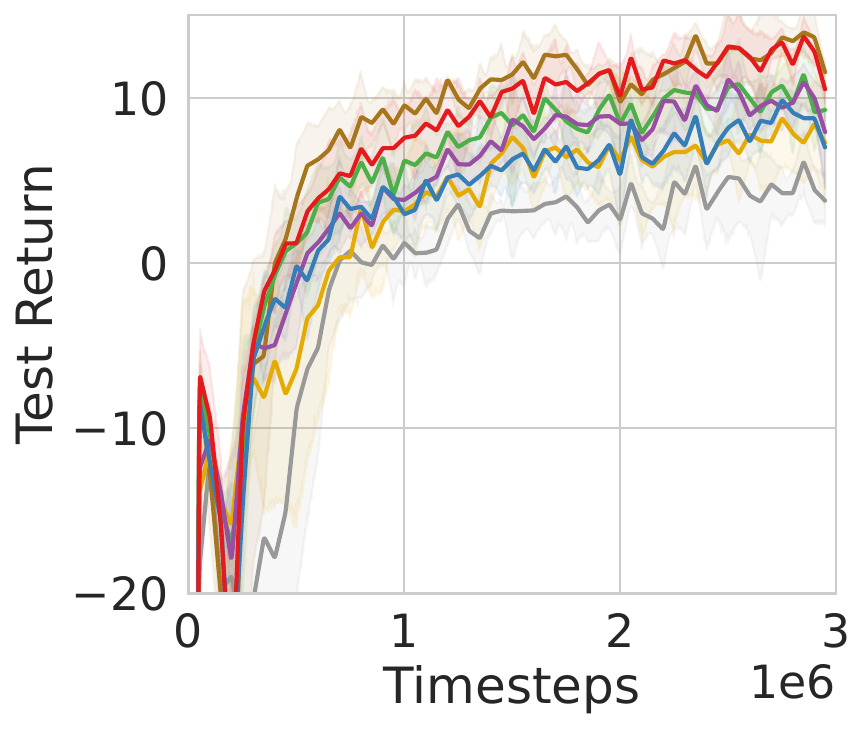}
    \caption{Push}
\end{subfigure}
\begin{subfigure}[t]{.24\textwidth}
    \centering
    \includegraphics[width=\textwidth]{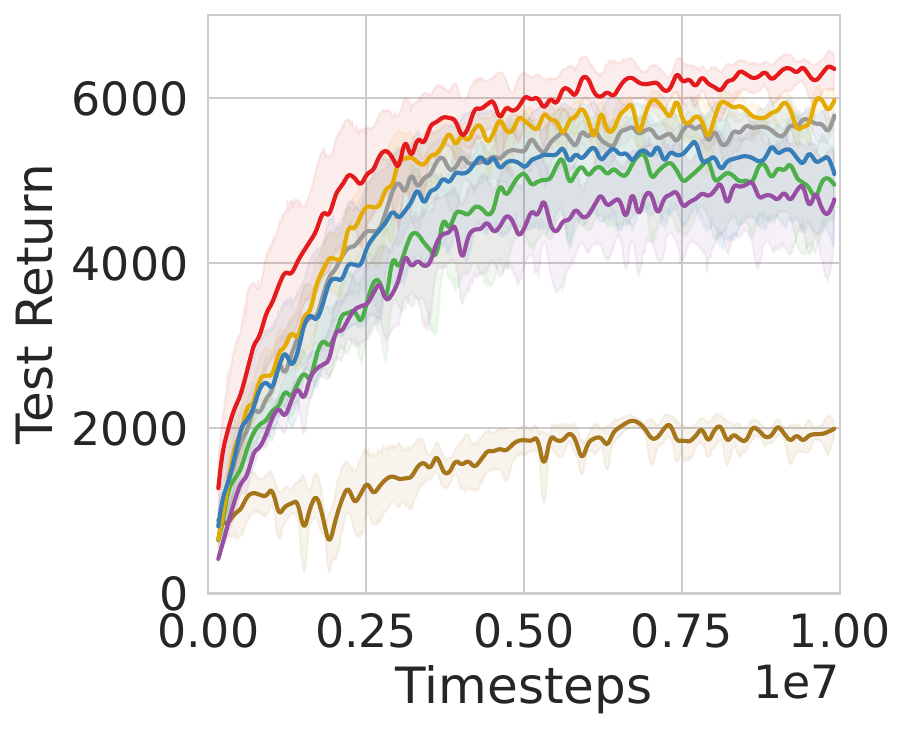}
    \caption{Ant-v2-4x2}
    \label{subfig: ant_4x2}
\end{subfigure}
\begin{subfigure}[t]{.24\textwidth}
    \centering
    \includegraphics[width=\textwidth]{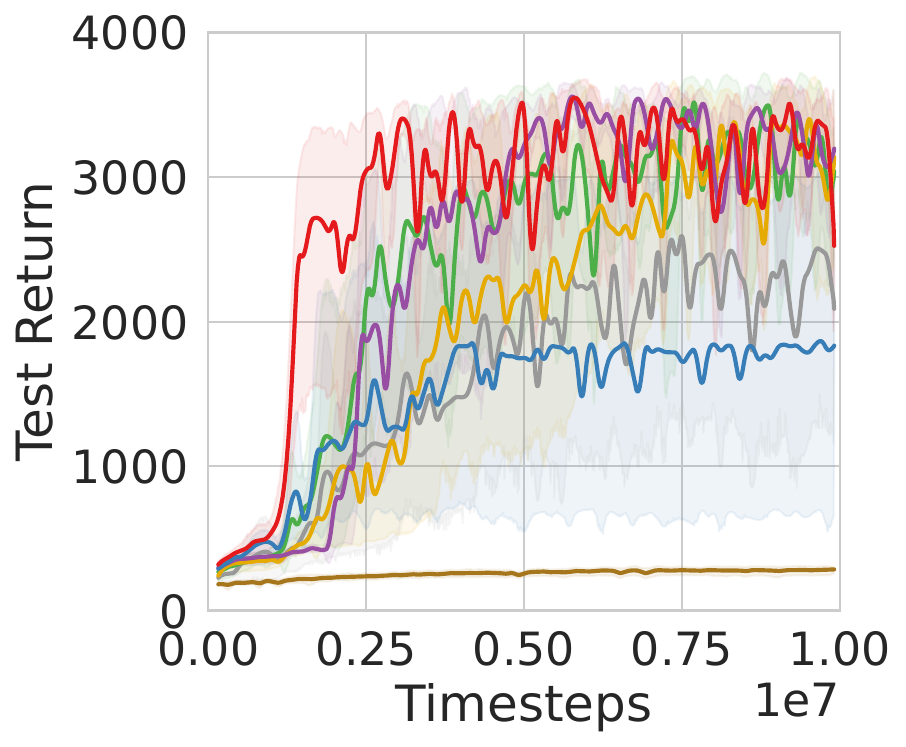}
    \caption{Hopper-v2-3x1}
\end{subfigure}
\begin{subfigure}[t]{.24\textwidth}
    \centering
    \includegraphics[width=\textwidth]{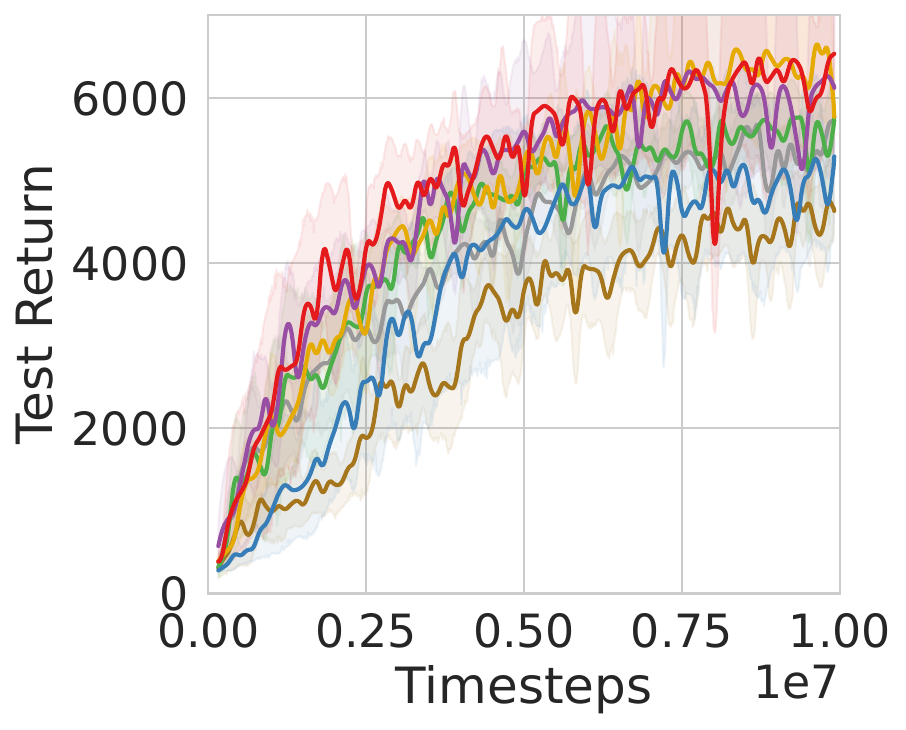}
    \caption{Walker2D-v2-2x3}
\end{subfigure}
\begin{subfigure}[t]{.24\textwidth}
    \centering
    \includegraphics[width=\textwidth]{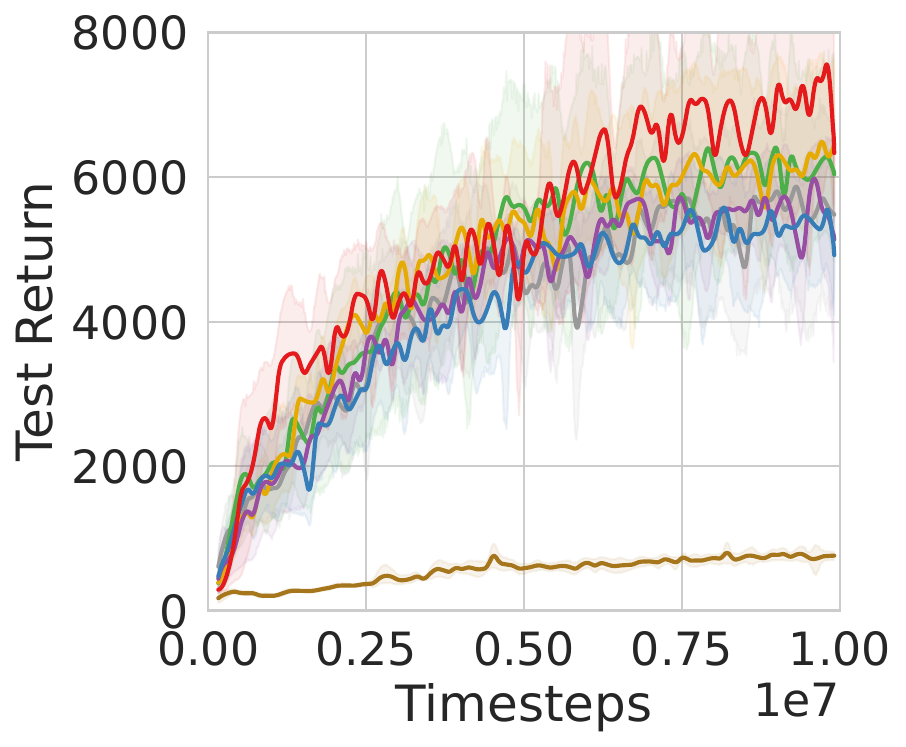}
    \caption{Walker2D-v2-6x1}
\end{subfigure}
\begin{subfigure}[t]{.24\textwidth}
    \centering
    \includegraphics[width=\textwidth]{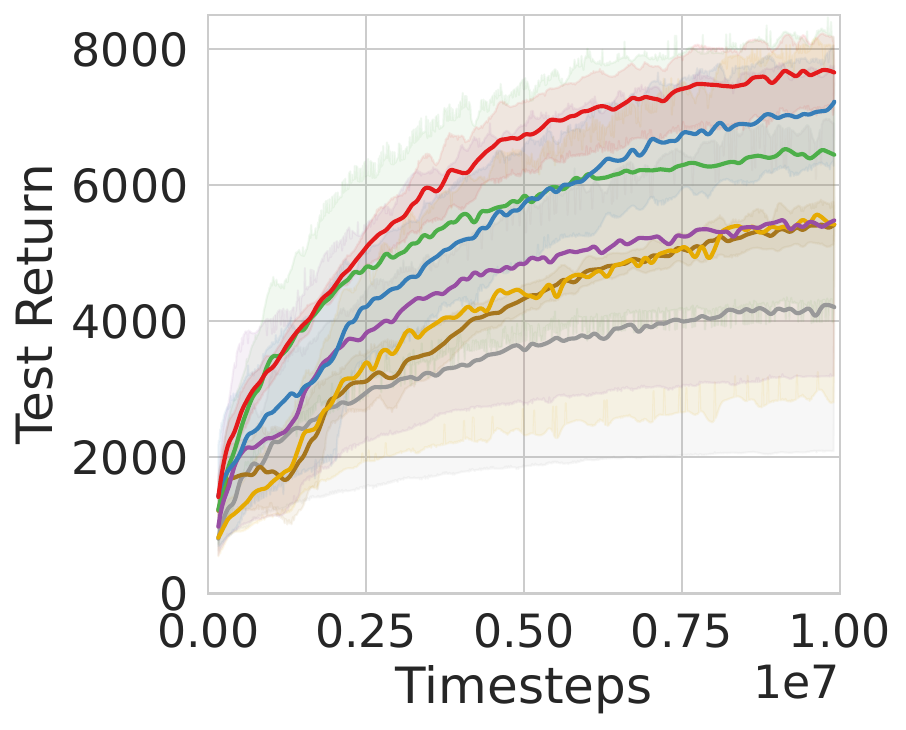}
    \caption{HalfCheetah-v2-2x3}
\end{subfigure}
\begin{subfigure}[t]{.24\textwidth}
    \centering
    \includegraphics[width=\textwidth]{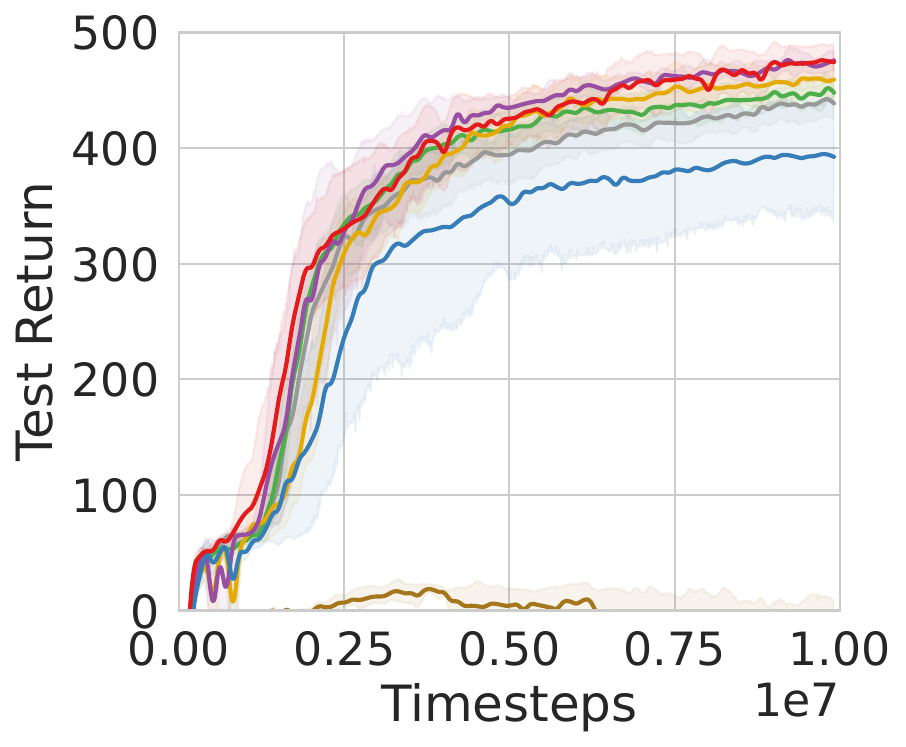}
    \caption{Swimmer-v2-10x2}
\end{subfigure}
\caption{Performance comparison with baselines on MPE and MaMuJoCo benchmarks.}
\label{fig:results_mpe_mamujoco}
\end{figure}

\begin{figure}[t]
\centering

\begin{subfigure}[c]{.26\textwidth}
    \centering
    \includegraphics[width=\textwidth]{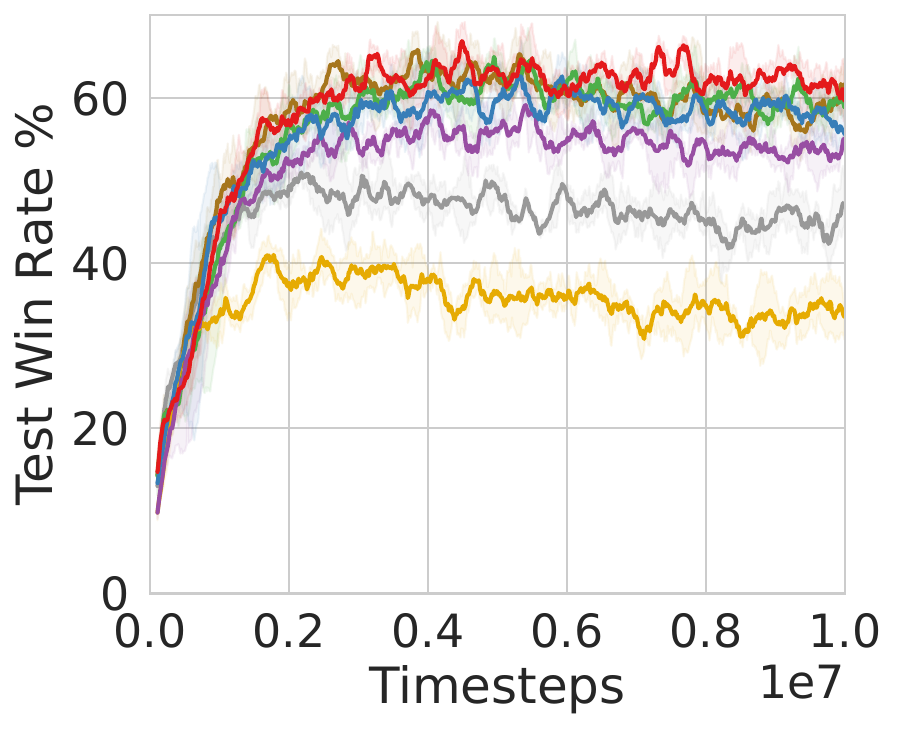}
    \caption{Terran\_5\_vs\_5}
\end{subfigure}
\begin{subfigure}[c]{.26\textwidth}
    \centering
    \includegraphics[width=\textwidth]{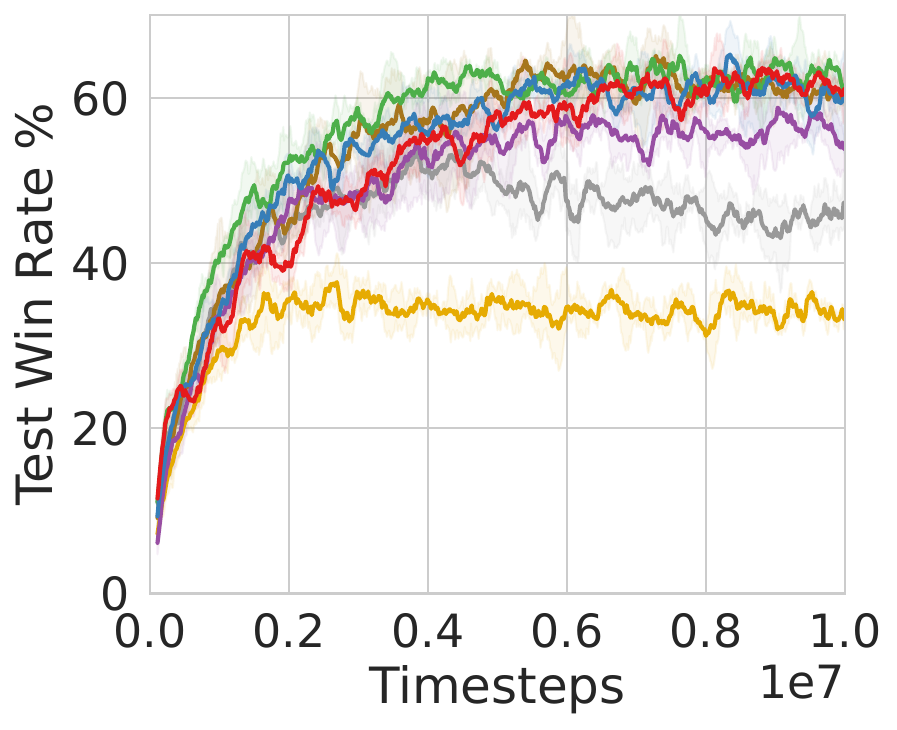}
    \caption{Protoss\_5\_vs\_5}
\end{subfigure}
\begin{subfigure}[c]{.26\textwidth}
    \centering
    \includegraphics[width=\textwidth]{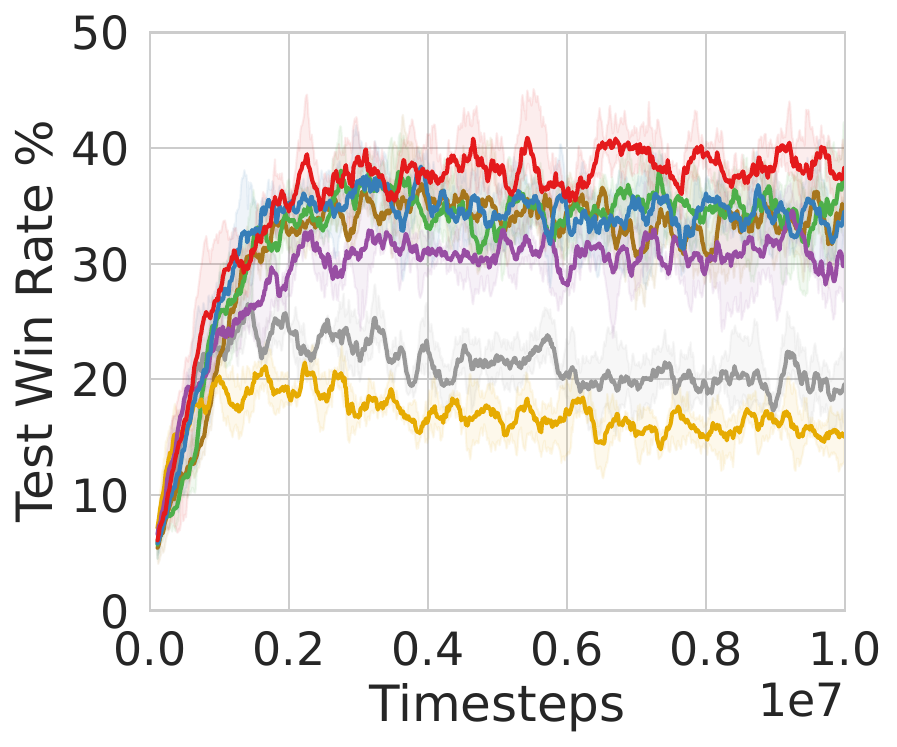}
    \caption{Zerg\_5\_vs\_5}
\end{subfigure}
\raisebox{0.5cm}{
\begin{subfigure}[c]{.16\textwidth}
    \centering
    \includegraphics[width=\textwidth]{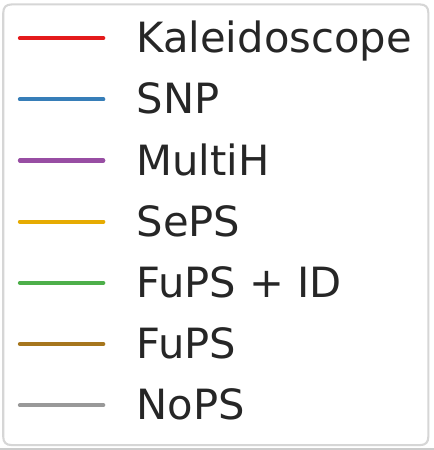}
\end{subfigure}}
\caption{Performance comparison with baselines on SMACv2 benchmarks.}
\label{fig:results_smacv2}
\end{figure}

\begin{figure}[t]
    \centering
    \begin{subfigure}[t]{.24\textwidth}
        \centering
        \includegraphics[width=\textwidth]{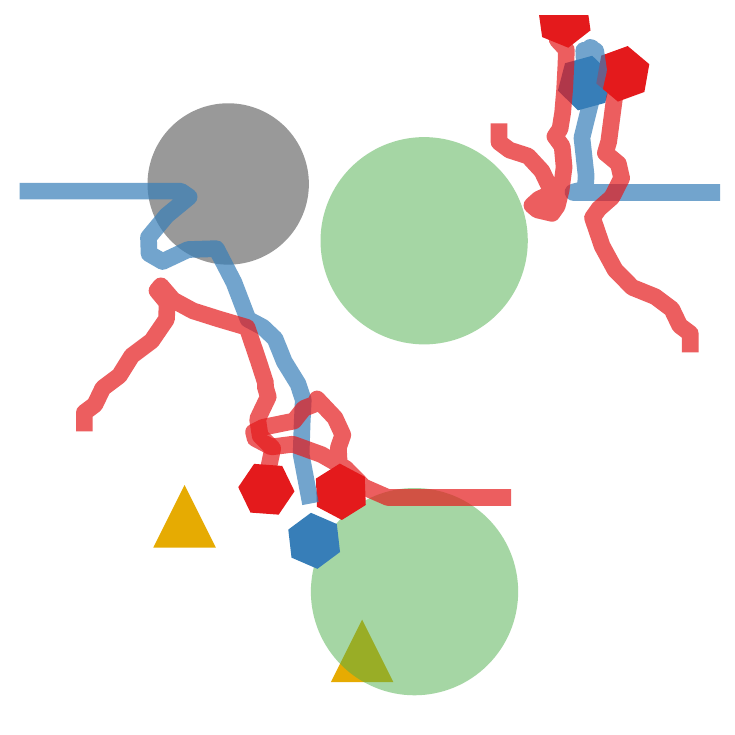}
        \caption{Trained policies.}
        \label{subfig: vis_policy}
    \end{subfigure}
    \begin{subfigure}[t]{.72\textwidth}
        \centering
        \includegraphics[width=\textwidth]{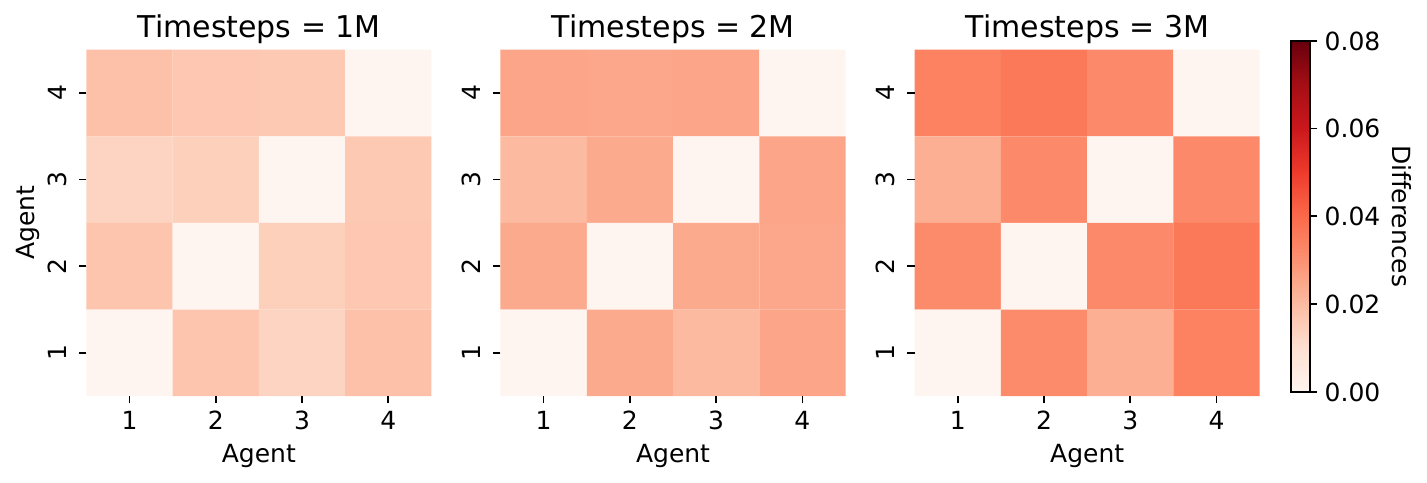}
        \caption{Pairwise mask differences among agents at different training timesteps.}
        \label{subfig: vis_masks}
    \end{subfigure}
    \caption{Visualization on \texttt{World}.}
    \label{fig: vis}
    \vspace{-12pt}
\end{figure}

We present the comparative performance of Kaleidoscope and baselines in \cref{fig:results_mpe_mamujoco} and \cref{fig:results_smacv2}. Overall, Kaleidoscope demonstrates superior performance, attributable to the flexibility of the learnable masks and the effectiveness of diversity regularization. 
Additionally, we observe that FuPS + ID generally outperforms NoPS, except for the \texttt{Ant-v2-4x2} scenario (\cref{subfig: ant_4x2}). This advantage is largely due to FuPS's higher sample efficiency; a single transition data sample updates the model parameters $N$ times in FuPS + ID, once for each agent, compared to just once in NoPS. Consequently, FuPS + ID models learn faster from the same number of transitions. Similarly, Kaleidoscope benefits from this mechanism as it shares weights among agents, allowing a single transition to update the model parameters multiple times. Furthermore, by integrating policy heterogeneity through learnable masks, Kaleidoscope enables diverse agent behaviors, as illustrated in the visualization results in \cref{fig: vis}. Ultimately, Kaleidoscope effectively balances parameter sharing and diversity, outperforming both full parameter sharing and non-parameter sharing approaches.  \looseness=-1

\paragraph{Cost analysis} Despite its superior performance, Kaleidoscope does not increase computational complexity at test time compared to the baselines.
We report the test time averaged FLOPs comparison of Kaleidoscope and baselines in \cref{tab: costs}. We see that due to the masking technique, Kaleidoscope has lower FLOPs compared to baselines, thereby enjoying a faster inference speed when being deployed. \looseness=-1

\begin{table}[th]
\caption{Averaged FLOPs (with calculation methods detailed in \cref{supp: flops_Cal}) across different methods. Results are first normalized with respect to the FuPS + ID model for each scenario and then averaged across scenarios within each environment (detailed results in \cref{supp: detailed_cost}). The lowest costs are highlighted in \textbf{bold}. \looseness=-1}
\label{tab: costs}
\begin{center}
\begin{tabular}{cccccccc}
\toprule
Methods & NoPS & FuPS & FuPS +ID & SePS & MultiH & SNP & Kaleidoscope \\
\midrule
MPE & 1.0x & 0.992x & 1.0x & 1.0x & 1.0x & 0.988x & \textbf{0.901x} \\
MaMuJoCo & 1.0x & 0.985x & 1.0x & 1.0x & 1.0x & 0.900x & \textbf{0.680x} \\
SMACv2 & 1.0x & 0.992x & 1.0x & 1.0x & 1.0x & 0.988x & \textbf{0.890x} \\
\bottomrule
\end{tabular}
\end{center}
\vspace{-10pt}
\end{table}

\paragraph{Ablation studies}


\begin{figure}[t]
    \centering
    \begin{minipage}{0.48\textwidth}
        \centering
        \begin{subfigure}[t]{0.49\textwidth}
            \centering
            \includegraphics[width=\textwidth]{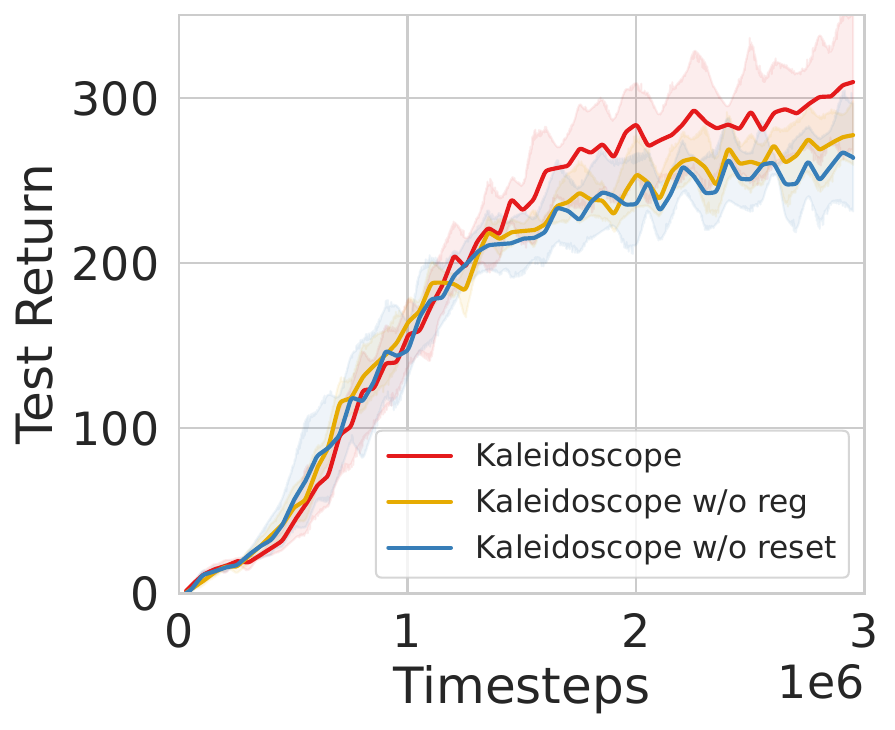}
            \caption{World}
            \label{subfig: abla_mpe}
        \end{subfigure}
        \begin{subfigure}[t]{0.49\textwidth}
            \centering
            \includegraphics[width=\textwidth]{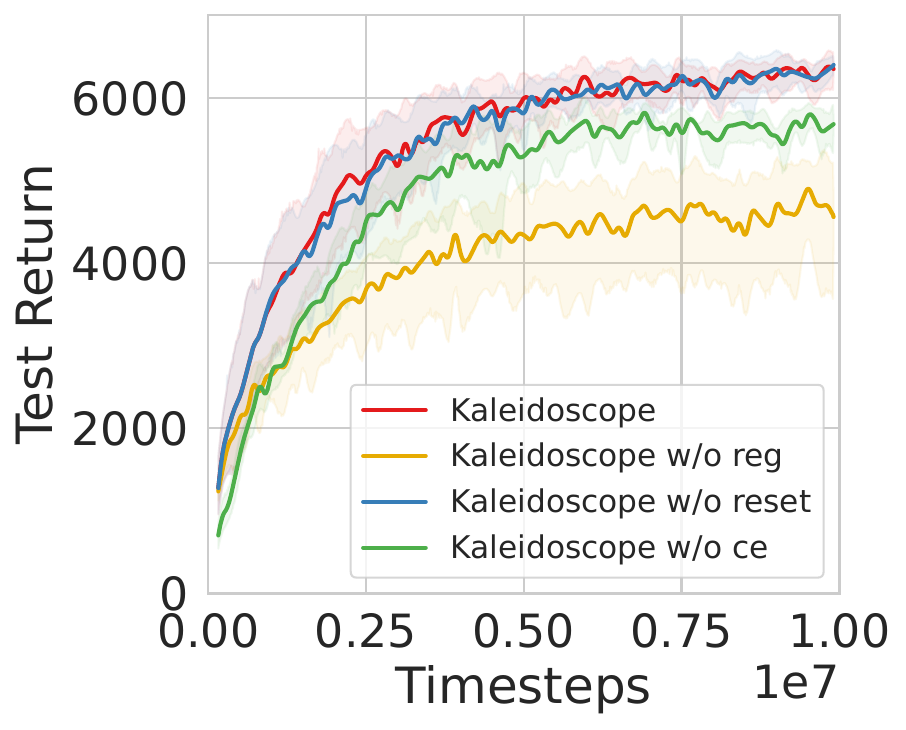}
            \caption{Ant-v2-4x2}
            \label{subfig: abla_mamujoco}
        \end{subfigure}
        \caption{Ablation studies.}
        \label{fig: ablation}
    \end{minipage}%
    \hfill
    \begin{minipage}{0.48\textwidth}
        \centering
        \begin{subfigure}[t]{0.49\textwidth}
            \centering
            \includegraphics[width=\textwidth]{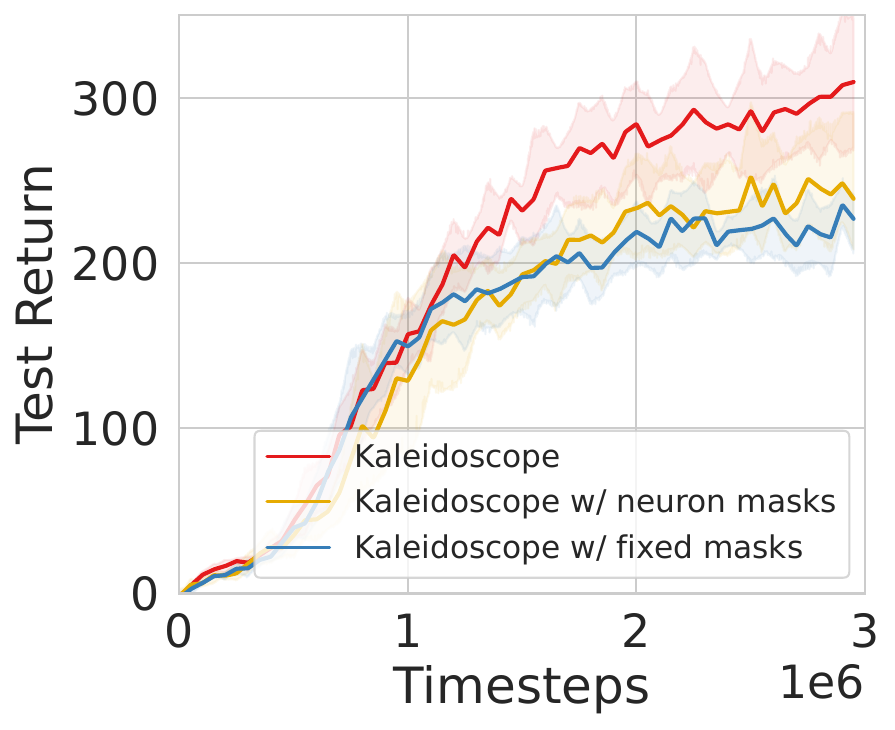}
            \caption{World}
            \label{subfig: comp_mpe}
        \end{subfigure}
        \begin{subfigure}[t]{0.49\textwidth}
            \centering
            \includegraphics[width=\textwidth]{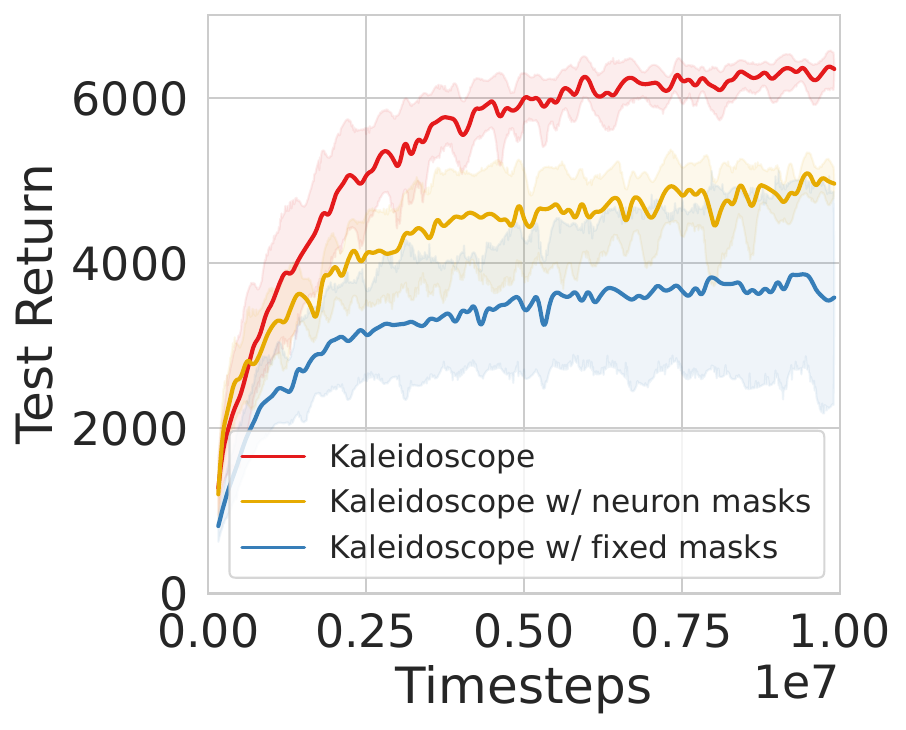}
            \caption{Ant-v2-4x2}
            \label{subfig: comp_mamujoco}
        \end{subfigure}
        \caption{Comparison on mask designs.}
        \label{fig: Comp_mask}
    \end{minipage}
    \vspace{-.1in}
\end{figure}
We conduct ablation studies to assess the impact of key components in Kaleidoscope, with results presented in \cref{fig: ablation}. Specifically, we compare Kaleidoscope with three ablations: 1) \textit{Kaleidoscope w/o reg}, which lacks the regularization term in \cref{eq: a_div} that encourages the masks to be distinct. 2) \textit{Kaleidoscope w/o reset}, which does not reset parameters. 3) \textit{Kaleidoscope w/o ce}, which does not use Kaleidoscope parameter sharing in critic ensembles and instead maintains two independent sets of parameters for critics. 
From the results, we observe that diversity regularization contributes the most to the performance of Kaleidoscope. Without it, masking degrades the performance due to the reduced number of parameters in each policy network. Resetting primarily aids learning in the late stages of training when needed, which aligns with the observation made by \citet{primacy_bias}. Notably, even with resetting, the performance does not experience abrupt drops thanks to the guidance provided by the masks on where to reset. When ablating the critic ensembles with Kaleidoscope parameter sharing, we observe inferior performance from the beginning of the training. This is because the critic ensembles with Kaleidoscope parameter sharing enable a higher UTD ratio of the critics, as discussed in \cref{sec: critics}. 

Furthermore, we conduct experiments to study the impact of mask designs. The results are shown in Figure \cref{fig: Comp_mask}. Specifically, we compare original Kaleidoscope with two alternative mask design choices: 1) \textit{Kaleidoscope w/ neuron masks}, where adaptive masking techniques are applied to neurons rather than weights. 2) \textit{Kaleidoscope w/ fixed masks}, where the masks are initialized at the beginning of training and kept fixed throughout the learning process. The results show that performance drops with either alternative design choice, demonstrating that Kaleidoscope's superior performance originates from the flexibility of the learnable masks on weights.

More results on hyperparameter analysis are included in \cref{supp: hyper_ana}.



\paragraph{Visualization}
We visualize the trained policies of Kaleidoscope on \texttt{World}, as shown in \cref{subfig: vis_policy}. The agents exhibit cooperative divide-and-conquer strategies (four red agents divide into two teams and surround the preys), contrasting with the homogeneous policies depicted in ~\cref{fig: intro_example}. 
We further examine the distinctions in the agents' masks and present the results in \cref{subfig: vis_masks}. First, we observe that by the end of the training, each agent has developed a unique mask, revealing that distinct masks facilitate diverse policies by selectively activating different segments of the neural network weights. Second, throughout the training process, we note that the differences among the agents' masks evolve dynamically. This observation confirms that Kaleidoscope effectively enables dynamic parameter sharing among the agents based on the learning progress, empowered by the adaptability of the learnable masks. More visualization results are provided in \cref{supp: further_vis}.

\section{Related Work}
\paragraph{Parameter sharing} 
First introduced by \citet{ps}, parameter sharing has been widely adopted in MARL algorithms~\citep{COMA,QMIX,MAPPO}, due to its simplicity and high sample efficiency~\citep{revisiting_ps}. However, schemes without parameter sharing typically offer greater flexibility for policy representation. To balance sample efficiency with policy representational capacity, some research efforts aim to find effective partial parameter sharing schemes. Notably, SePS~\citep{SePS} first clusters agents based on their transitions at the start of training and restricts parameter sharing within these clusters. Subsequently, SNP~\citep{snp} enables partial parameter sharing by utilizing the lottery ticket hypothesis~\citep{LTH} to initialize heterogeneous network structures. Concurrent to our work, AdaPS~\citep{AdaPS} combines SNP and SePS by proposing a cluster-based partial parameter sharing scheme.
While these methods have shown promise in certain domains, their performance potential is often limited by the static nature of the parameter sharing schemes set early in training. 
Our proposed Kaleidoscope distinguishes itself by dynamically learning specific parameter sharing configurations alongside the development of MARL policies, thereby offering enhanced training flexibility.

\paragraph{Agent heterogeneity in MARL}
To incorporate agent heterogeneity in MARL and enable diverse behaviors among agents, previous methods have explored concepts such as diversity and roles. Specifically, diversity-based approaches aim to enhance pairwise distinguishability among agents based on identities~\citep{EOI}, trajectories~\citep{CDS}, or credits assignment~\citep{CIA,ACORM} through contrastive learning techniques. Concurrently, role-based strategies, sometimes referred to as skills~\citep{skill_obs} or subtasks~\citep{MACC}, employ conditional policies to differentiate agents by assigning them to various conditions. These conditions may be based on agent identities~\citep{LDSA}, local observations~\citep{skill_obs}, local histories~\citep{ROMA,RODE, MACC} or joint histories~\citep{COPA,ALMA,SIRD}.
This line of researches mainly focus on module design and operate separately from parameter-level adjustments, making them orthogonal to our approach. Nevertheless, integrating these methods with our work could potentially enhance performance further.

\paragraph{Sparse networks in deep reinforcement learning (RL)}
Although relatively few, there are some noteworthy recent attempts to find sparse networks for deep RL. In particular, PoPS~\citep{PoPS} prunes the dense networks post-training, achieving significantly reduced execution time complexity. Additionally, \citep{LTH_RL} validate the lottery ticket hypothesis within the RL domain, producing high-performance models even under extreme pruning rates. Subsequent efforts, including DST~\citep{DST}, TE-RL*~\citep{TE_RL} and RLx2~\citep{RLx2} employ topology evolution (TE) techniques to further decrease the training costs.
While these developments utilize sparse training techniques, which are similar to the methods we employ, their primary focus is on reducing training and execution costs in single-agent settings. In contrast, our work leverages sparse network strategies as a means to enhance parameter sharing techniques, aiming to improve MARL performance.


\section{Conclusions and Future Work}

In this work, we introduced \emph{Kaleidoscope}, a novel adaptive partial parameter sharing mechanism for MARL. It leverages distinct learnable masks to facilitate network heterogeneity, applicable to both agent policies and critic ensembles. Specifically, Kaleidoscope is built on three technical components: STR-empowered learnable masks, network diversity regularization, and a periodic resetting mechanism. When applied to agent policy networks, Kaleidoscope balances sample efficiency and network representational capacities. In the context of critic ensembles, it improves value estimations. By combining our proposed Kaleidoscope with QMIX and MATD3, we have empirically demonstrated its effectiveness across various MARL benchmarks. This study shows great promises in developing adaptive partial parameter sharing mechanisms to enhance the performance of MARL. For future work, it is interesting to further extend Kaleidoscope to other domains such as offline MARL or meta-RL.

\section*{Acknowledgements}
This work was supported by the Hong Kong Research Grants Council under the NSFC/RGC Collaborative Research Scheme grant CRS\_HKUST603/22.
And we thank the anonymous reviewers for their valuable feedback and suggestions.

\bibliographystyle{ACM-Reference-Format} 
\bibliography{main.bib}

\newpage
\appendix

\section{Experimental details}

\subsection{Implementation details} \label{supp: implementation_details}

\subsubsection{Kaleidoscope with MATD3} \label{suppl: training_obj_MATD3}

\paragraph{Critic ensembles}
When incorporating Kaleidoscope into the MATD3 algorithm, the overall training loss for the critic ensembles becomes: 
\begin{equation}
    \mathcal{L}_c^{\text{all}}(\phi_0, \bm{s}^c) = \sum_{j = 1, \ldots, K} \mathcal{L}_c^{\text{all}}(\phi_j) = \sum_{j = 1, \ldots, K} \mathcal{L}_c(\phi_0, \bm{s}^c_j) - \alpha^{d} \cdot \mathcal{J}^{\text{div}}_c(\bm{s}^c),
    \label{eq: supp_overall_c_obj}
\end{equation}
with $\mathcal{L}_c(\phi_0, \bm{s}^c_j)$ being the original MATD3 loss given in \cref{eq: c_j_obj}, $\mathcal{J}^{\text{div}}_c(\bm{s}^c)$ being the diversity regularization given in \cref{eq: c_div}, and $\alpha^{d}$ being a coefficient balancing the original MARL objective and the proposed diversity regularization. Note that although $\mathcal{J}^{\text{div}}_c(\bm{s}^c)$ contains parameters $\phi_0$, we stop the gradients for $\phi_0$ in $\mathcal{J}^{\text{div}}_c(\bm{s}^c)$. 

In the implementation, we apply layer-wise weights to the diversity regularization term $\mathcal{J}^{\text{div}}_c(\bm{s}^c)$, which is defined as
\begin{equation}
    \mathcal{J}^{\text{div}}_c(\bm{s}^c) = 
    \sum_{l=1, \ldots, L} w_l \cdot
    \sum_{i = 1, \ldots, K} 
    \sum_{\substack{{j = 1, \ldots, K}\\ j\neq i}}  \lVert  \phi_0 \odot (\bm{M}_{i, l}^c - \bm{M}_{j, l}^c) \rVert_1,
    \label{eq: c_div_layer_wise}
\end{equation}
where $l$ denotes the layer index of the neuron networks, $L$ represents the total number of layers, $\bm{M}_{i, l}^c$ is the mask for agent $i$ at layer $l$, and the layer-wise weights are set as $w_l = 2^l$. The intuition behind this choice is that features closer to the output tend to be more compact~\citep{STR}; consequently, assigning larger regularization weights to these layers may have a more significant impact on the output action decisions. Our initial experiments empirically demonstrate that setting $w_l = 2^l$ improves performance compared to the case where $w_l = 1$. Based on these findings, we maintain this design choice throughout all our experiments.

In practice, we adaptively adjust $\alpha^{d}$ while maintaining a constant ratio between the MATD3 loss and the diversity loss, which is treated as a hyperparameter:
\begin{equation}
    \alpha^{d} = \frac{\lvert  \sum_{j = 1, \ldots, K} \mathcal{L}_c(\phi_0, \bm{s}^c_j) \rvert}{\lvert \mathcal{J}^{\text{div}}_c(\bm{s}^c) \rvert} \cdot \alpha,
    \label{eq: suppl_c_coefficient}
\end{equation}
where $\alpha$ is a hyperparameter, and the gradients for $\frac{\lvert  \sum_{j = 1, \ldots, K} \mathcal{L}_c(\phi_0, \bm{s}^c_j) \rvert}{\lvert \mathcal{J}^{\text{div}}_c(\bm{s}^c) \rvert}$ are stopped.

\paragraph{Actors} For the actors, the training objective is to maximize the following term
\begin{equation}
    \mathcal{J}^{\text{all}}(\theta_0, \bm{s}) = \sum_{i = 1, \ldots, n} \mathcal{J}^{\text{all}}(\theta_i) = \sum_{i = 1, \ldots, n} \mathcal{J}(\theta_0, \bm{s}_i) + \beta^d \cdot \mathcal{J}^{\text{div}}(\bm{s}),
\end{equation}
where $\mathcal{J}(\theta_0, \bm{s}_i)$ is the original actor objective defined in \cref{eq: matd3_a_obj}, $\mathcal{J}^{\text{div}}(\bm{s})$ is the diversity regularization given in \cref{eq: a_div} with layer-wise weights as in \cref{eq: c_div_layer_wise}, $\beta^d$ is the regularization coefficient. The value of $\beta^{d}$ is determined by 
\begin{equation}
    \beta^{d} = \frac{\lvert  \sum_{i = 1, \ldots, n} \mathcal{J}(\theta_0, \bm{s}_i) \rvert}{\lvert \mathcal{J}^{\text{div}}(\bm{s}) \rvert} \cdot \beta, 
\end{equation}
where $\beta$ is a constant hyperparameter, similar to the approach used in \cref{eq: suppl_c_coefficient} for the critic ensembles.

\subsubsection{Kaleidoscope with QMIX} \label{suppl: training_obj_QMIX}
When incorporating Kaleidoscope into the QMIX algorithm~\citep{QMIX}, we apply Kaleidoscope parameter sharing only to the local Q networks. Consequently, the training loss is defined as:
\begin{equation}
    \mathcal{L}^{\text{all}}(\theta_0, \bm{s}) =  \mathcal{L}(\theta_0, \bm{s}) - \beta^d \cdot \mathcal{J}^{\text{div}}(\bm{s}),
\end{equation}
where 
\begin{equation}
    \mathcal{L}(\theta_0, \bm{s}) = \mathbb{E}_{(s^t, \bm{o}^t,  \bm{a}^t, r^t, s^{t+1}, \bm{o}^{t+1}) \sim \mathcal{D}} \left[\left(y^{tot} - Q_{tot}(s^t, \boldsymbol{o}^t, \boldsymbol{a}^t; \theta_0, \bm{s}) \right)^2\right],
\end{equation}
with $y^{tot} = r + \gamma \max_{\boldsymbol{a}} Q_{tot}(s^{t+1}, \boldsymbol{o}^{t+1}, \boldsymbol{a}; \theta^-)$ and $\theta^-$ representing the parameters of a target network as in DQN.

\subsubsection{Network architecture and hyperparameters}\label{suppl： suppl_network_arch_hyper}

\paragraph{Codebase} 
Our implementation of Kaleidoscope and baseline algorithms are based on the following codebase:
\begin{itemize}
    \item HARL~\citep{harl} (MATD3 implementation): \url{https://github.com/PKU-MARL/HARL}
    \item EPyMARL~\citep{epymarl} (QMIX implementation for MPE): \\\url{https://github.com/uoe-agents/epymarl}
    \item PyMARL2~\citep{pymarl2} (QMIX implementation for SMACv2): \\\url{https://github.com/benellis3/pymarl2}
    \item SePS~\citep{SePS}: \url{https://github.com/uoe-agents/seps}
\end{itemize}

The code for Kaleidoscope is publicly available at \url{https://github.com/LXXXXR/Kaleidoscope}. 

\paragraph{Network architecture} In line with prior works \citep{harl, epymarl, pymarl2}, we employ deep neural networks consisting of multilayer perceptrons (MLPs) with rectified linear unit (ReLU) activation functions and gated recurrent units (GRUs) to parameterize the actor and critic networks. Moreover, when the masking technique is applied to critic ensembles, we incorporate layer normalization between the MLP layers and ReLU activations \citep{dropout_q}. In Kaleidoscope, the masking technique is applied to the MLP layers, and the resetting mechanisms described in \cref{sec: reset,sec: critics} are applied to the last three layers of the respective neural networks, following \citet{primacy_bias}.

\paragraph{Hyperparameters}

To ensure a fair comparison, we implement our method and all the baselines using the same codebase with the same set of hyperparameters, with the exception of method-specific ones. The common hyperparameters are listed in  \cref{tab: hyper_mamujoco,tab: hyper_mpe,tab: hyper_smacv2}. The Kaleidoscope-specific hyperparameters are provided in \cref{tab: hyper_Kalei}.

\begin{table}[th]
    \centering
    \caption{Common hyperparameters used for MATD3 in the MaMuJoCo domain.}
    \label{tab: hyper_mamujoco}
    \begin{threeparttable}
    \begin{tabular}{cc}
    \toprule
    Hyperparameter & Value \\
    \midrule
    Number of layers  & $3$\\
    Hidden sizes  & $256$\\
    Discount factor $\gamma$ & $0.99$ \\
    Rollout threads & $10$ \\
    Critic lr & $1 \times 10^{-3}$ \\
    Actor lr & $5 \times 10^{-4}$ \\
    Exploration noise & $0.1$ \\
    Batch size & $1000$ \\
    Replay buffer size & $1 \times 10^{6}$ \\
    Number of environment steps & $10 \times 10^{6}$ \\
    n\_step\tnote{1}  & $(5, 10, 20)$ \\
    \bottomrule
    \end{tabular}
    \begin{tablenotes}
        \item[1] Here we adopt the per-scenario finetuned value for this hyperparameter as provided by HARL.
    \end{tablenotes}
    \end{threeparttable}
\end{table}

\begin{table}[th]
    \centering
    \caption{Common hyperparameters used for QMIX in the MPE domain.}
    \label{tab: hyper_mpe}
    \begin{tabular}{cc}
    \toprule
    Hyperparameter & Value \\
    \midrule
    Number of layers  & $5$\\
    Hidden sizes  & $64$\\
    Discount factor $\gamma$ & $0.99$ \\
    Lr & $5 \times 10^{-4}$ \\
    Initial $\epsilon$ & $1.0$ \\
    Final $\epsilon$ & $0.05$ \\
    Batch size & $32$ \\
    Replay buffer size & $5000$ \\
    Number of environment steps & $3 \times 10^{6}$ \\
    Double Q & True \\
    \bottomrule
    \end{tabular}
\end{table}

\begin{table}[th]
    \centering
    \caption{Common hyperparameters used for QMIX in the SMACv2 domain.}
    \label{tab: hyper_smacv2}
    \begin{tabular}{cc}
    \toprule
    Hyperparameter & Value \\
    \midrule
    Number of layers  & $5$\\
    Hidden sizes  & $64$\\
    Discount factor $\gamma$ & $0.99$ \\
    Lr & $1 \times 10^{-3}$ \\
    Initial $\epsilon$ & $1.0$ \\
    Final $\epsilon$ & $0.05$ \\
    Batch size & $128$ \\
    Replay buffer size & $5000$ \\
    Number of environment steps & $10 \times 10^{6}$ \\
    Double Q & False \\
    \bottomrule
    \end{tabular}
\end{table}

\begin{table}[th]
    \centering
    \caption{Hyperparameters used for Kaleidoscope. }
    \label{tab: hyper_Kalei}
    \begin{tabular}{ccc}
    \toprule
    Hyperparameter & Environment & Value \\
    \midrule
    \multirow{3}{*}{Actor diversity coefficient $\beta$} & MaMuJoCo & $0.1$ \\ 
    & MPE & $0.5$ \\
    & SMACv2 & $5.0$ \\
    \hline \noalign{\vskip 2pt}
    \multirow{3}{*}{Actors reset probability $\rho$} & MaMuJoCo & $0.5$ \\
    & MPE & $0.1$ \\
    & SMACv2 & $0.2$ \\
    \hline \noalign{\vskip 2pt}
    \multirow{3}{*}{Actor reset interval} & MaMuJoCo & $1 \times 10^{6}$ \\
    & MPE & $200 \times 10^{3}$ \\
    & SMACv2 & $1 \times 10^{6}$  \\
    \hline \noalign{\vskip 2pt}
    Number of critic ensembles $K$ & MaMuJoCo & $5$ \\
    Critic ensembles diversity coefficient $\alpha$ & MaMuJoCo & $0.1$ \\
    Critic reset interval & MaMuJoCo & $800 \times 10^{3}$ \\
    \bottomrule
    \end{tabular}
\end{table}

\subsection{Environmental details} \label{supp: env_details}

\paragraph{Codebase}
The environments used in this work are listed below with descriptions in \cref{tab: env}.
\begin{itemize}
    \item MaMuJoCo~\citep{facmac_mamujoco}: \url{https://github.com/schroederdewitt/multiagent_mujoco}
    \item MPE~\citep{MADDPG,epymarl}: \url{https://github.com/semitable/multiagent-particle-envs}
    \item SMACv2~\citep{smacv2}: \url{https://github.com/oxwhirl/smacv2}

\end{itemize}

\begin{table}[th]
    \centering
    \caption{Environments details.}
    \label{tab: env}
    \begin{tabular}{ccccc}
    \toprule
    Environment & Action space & Agent types & Scenarios & Number of agents \\
    \midrule
    \multirow{6}{*}{MaMuJoCo} & \multirow{6}{*}{Continuous} & \multirow{6}{*}{Heterogeneous, fixed} & Ant-v2-4x2 & $4$ \\
    & & & Hopper-v2-3x1 & $3$ \\
    & & & Walker2D-v2-2x3 & $2$ \\
    & & & Walker2D-v2-6x1 & $6$ \\
    & & & HalfCheetah-v2-2x3 & $2$ \\
    & & & Swimmer-v2-10x2 & $10$ \\
    \hline \noalign{\vskip 2pt}
    \multirow{2}{*}{MPE}  & \multirow{2}{*}{Discrete} & \multirow{2}{*}{Homogeneous, fixed} & World & $4$\\
    & & & Push & $5$ \\
    \hline \noalign{\vskip 2pt}
    \multirow{3}{*}{SMACv2}  & \multirow{3}{*}{Discrete} & \multirow{3}{*}{Heterogeneous, dynamic} & Terran\_5\_vs\_5 & $5$\\
    & & & Protoss\_5\_vs\_5 & $5$ \\
    & & & Zerg\_5\_vs\_5 & $5$ \\
    \bottomrule
    \end{tabular}
\end{table}

\paragraph{MPE} We extend the scenario settings provided in the original codebase to increase the complexity and challenge of the tasks. In \texttt{World}, we set the number of predators (agents) to 4, the number of prey to 2, the number of obstacles to 1, and the number of forests to 2. The objective of the game is for the predators to approach the prey while avoiding collisions with obstacles. The prey is attracted to the food and can hide from the predators in the forests. In \texttt{Push}, we set the number of agents to 5, the number of adversaries to 2, and the number of landmarks to 2. The goal of the game is for the agents to push the adversaries away from the landmarks.
In both scenarios, we pretrain the adversary (prey) policies using the MADDPG algorithm~\cite{MADDPG} and use these pretrained policies to test the performance of different algorithms

\subsection{FLOPs calculation} \label{supp: flops_Cal}
To calculate the number of floating-point operations (FLOPs) for a single forward pass of a sparse model, we sum the total number of multiplications and additions layer by layer, following the approach in \cite{RigL}. For a fully-connected layer, the FLOPs are computed as follows:
\begin{equation}
    \text{FLOPs} = 2 \times (1 - \text{Sparsity}) \times \text{In} \times \text{Out}.
\end{equation}

For a GRU cell, the FLOPs are computed as:
\begin{equation}
    \text{FLOPs} = 2 \times (3 \times \text{Hidden}^2 + 3 \times \text{In} \times \text{Hidden} +  13 \times \text{Hidden}) .
\end{equation}

\subsection{Experimental Infrastructure} \label{supp: infrast}
The experiments on the SMACv2 benchmark were conducted using NVIDIA GeForce RTX 3090 GPUs, while the experiments on other benchmarks were performed using NVIDIA GeForce RTX 3080 GPUs. Each experimental run required less than 2 days to complete.

\section{More results and discussion}
\subsection{Detailed Costs} \label{supp: detailed_cost}
We provide per-scenario FLOPs across different methods in \cref{tab: costs_all} as a supplement for \cref{tab: costs}. 



\begin{table}[th]
\caption{Averaged FLOPs for different methods. Results are normalized w.r.t. the FuPS + ID model. The lowest costs are highlighted in \textbf{bold}.}
\label{tab: costs_all}
\begin{center}
\begin{tabular}{lcccccccc}
\toprule
Scenarios & NoPS & FuPS &  FuPS +ID & SePS & MultiH & SNP & Kaleidoscope  \\
\midrule
World & 1.0x & 0.993x & 1.0x & 1.0x & 1.0x & 0.988x & \textbf{0.897x} \\
Push & 1.0x & 0.991x & 1.0x & 1.0x & 1.0x & 0.988x & \textbf{0.904x} \\ \midrule
Ant-v2-4x2 & 1.0x & 0.990x & 1.0x & 1.0x & 1.0x & 0.900x & \textbf{0.640x} \\
Hopper-v2-3x1 & 1.0x & 0.989x & 1.0x & 1.0x & 1.0x & 0.900x & \textbf{0.721x} \\
Walker2D-v2-2x3 & 1.0x & 0.992x & 1.0x & 1.0x & 1.0x & 0.900x & \textbf{0.731x} \\
Walker2D-v2-6x1 & 1.0x & 0.979x & 1.0x & 1.0x & 1.0x & 0.900x & \textbf{0.763x} \\
HalfCheetah-v2-2x3 & 1.0x & 0.993x & 1.0x & 1.0x & 1.0x & 0.900x & \textbf{0.614x} \\
Swimmer-v2-10x2 & 1.0x & 0.968x & 1.0x & 1.0x & 1.0x & 0.900x & \textbf{0.611x} \\ \midrule
Terran\_5\_vs\_5 & 1.0x & 0.992x & 1.0x & 1.0x & 1.0x & 0.988x & \textbf{0.890x} \\
Protoss\_5\_vs\_5 & 1.0x & 0.992x & 1.0x & 1.0x & 1.0x & 0.988x & \textbf{0.895x} \\
Zerg\_5\_vs\_5 & 1.0x & 0.992x & 1.0x & 1.0x & 1.0x & 0.988x & \textbf{0.885x} \\
\bottomrule
\end{tabular}
\end{center}
\end{table}

\subsection{Hyperparameter Analysis} \label{supp: hyper_ana}

\begin{figure}
    \centering
    \begin{subfigure}[t]{.4\textwidth}
        \centering
        \includegraphics[width=\textwidth]{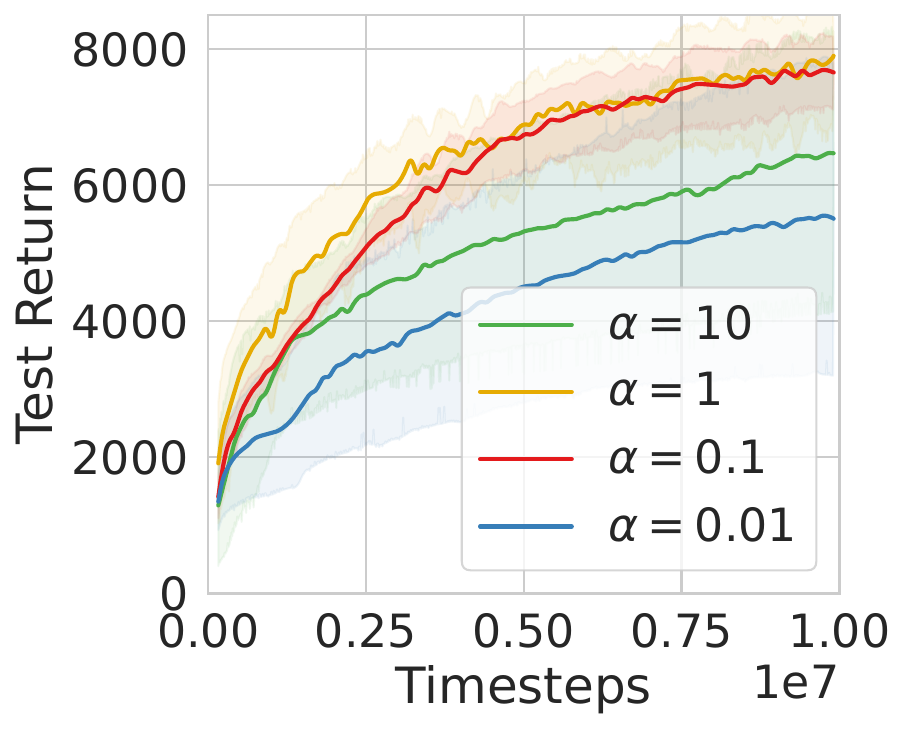}
        \caption{Critic ensembles diversity coefficient $\alpha$.}
        \label{subfig: alpha_ana}
    \end{subfigure}
    \begin{subfigure}[t]{.4\textwidth}
        \centering
        \includegraphics[width=\textwidth]{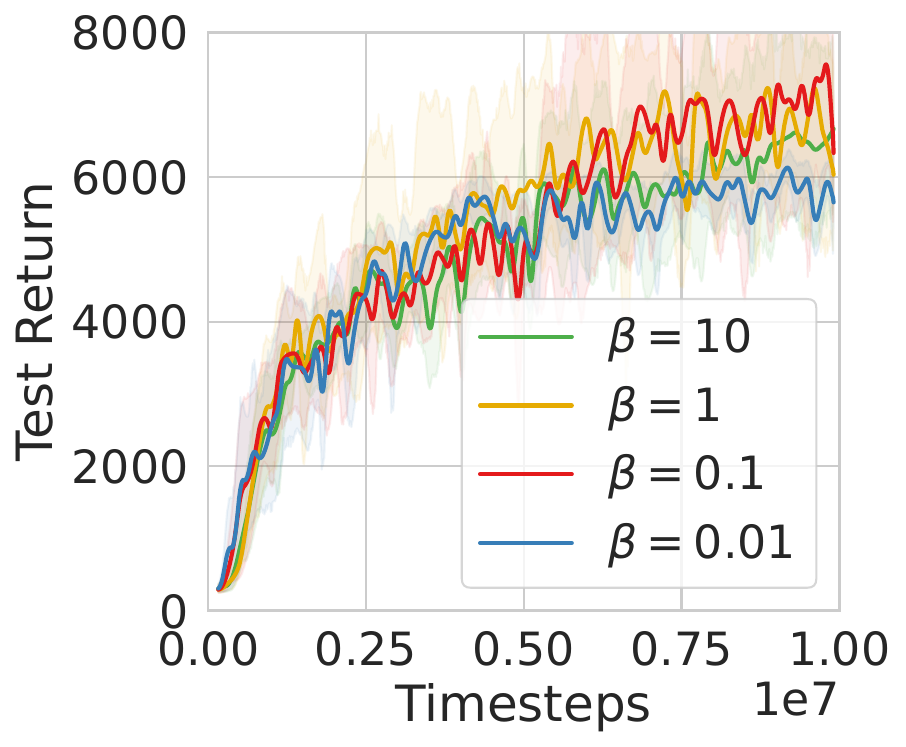}
        \caption{Actors diversify coefficient $\beta$.}
        \label{subfig: beta_ana}
    \end{subfigure}
    \caption{Hyperparameter analysis.}
    \label{fig: ana}
\end{figure}

We conduct further analysis on the hyperparameters $\alpha$ and $\beta$, and present the results in \cref{fig: ana}. The hyperparameter $\alpha$ controls the variance of the critic ensembles. As shown in \cref{subfig: alpha_ana}, we observe that an excessively small $\alpha$ results in degraded performance because it reduces the critic ensembles to a single critic network, causing the value estimation to suffer from severe overestimation. Conversely, an excessively large $\alpha$ also deteriorates performance, possibly due to increased estimation bias.
For $\beta$, as illustrated in \cref{subfig: beta_ana}, an overly small $\beta$ leads to degraded performance because it reduces the Kaleidoscope parameter sharing to full parameter sharing, confining the policies to be identical. An overly large $\beta$ also negatively impacts performance, as it may cause the training objective to deviate too much from minimizing the original MARL loss.

In general, we recommend setting both hyperparameters between $0.1$ and $1$. However, the optimal hyperparameter values may vary across different scenarios. For fair comparisons, we maintain the same set of hyperparameters across all scenarios in our experiments. Nevertheless, further tuning of these hyperparameters has the potential to enhance performance.

\subsection{Further Visualization Results} \label{supp: further_vis}

\begin{figure}
  \centering
  \includegraphics[width=0.8\linewidth]{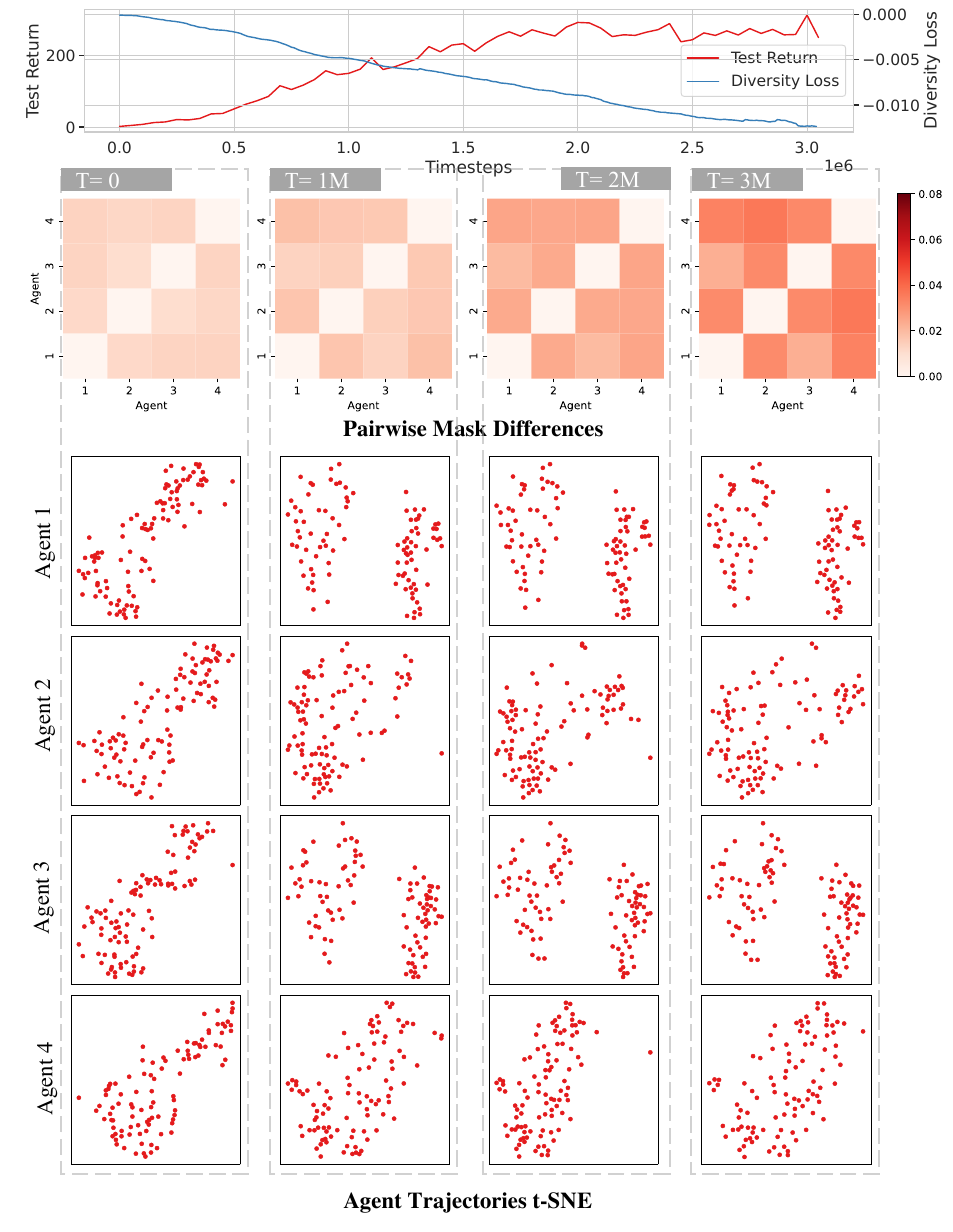}
  \caption{Further visualization on \texttt{World}.  \looseness=-1}
  \label{fig: more_vis}
\end{figure}

To better understand how learnable masks in Kaleidoscope affect the performance through policies, we visualize the pairwise mask differences among agents and the agent trajectories at different training stages in \cref{fig: more_vis}. As training progresses, the test return increases and diversity loss decreases, indicating better performance and greater diversity among agent policies. Correspondingly, mask differences among agents increase, and the agent trajectory distribution becomes more diverse.

\subsection{Limitations} \label{supp: limitations}
Here we discuss some limitations of Kaleidoscope.

First, as suggested by results in \cref{supp: hyper_ana}, the optimal hyperparameters vary from scenario to scenario. Therefore, using the same hyperparameters across all scenarios may not yield the best performance for Kaleidoscope. Developing an automatic scheme that utilizes environmental information to determine these hyperparameters would be beneficial. 

Second, as the environments used in this work contain no more than 10 agents, we assign a distinct mask for each agent. However, when the problem scales to hundreds of agents, this vanilla implementation may fail. 
In such cases, a possible approach is to cluster $N$ agents into $K$ ($K < N$) groups and train $K$ masks with Kaleidoscope. This would reduce computational costs and achieve a better trade-off between sample efficiency and diversity. Within the same group, agents share all parameters, while agents from different groups share only partial parameters. Techniques for clustering agents based on experience, as proposed by \cite{SePS}, could be useful.



\newpage
\section*{NeurIPS Paper Checklist}

\begin{enumerate}

\item {\bf Claims}
    \item[] Question: Do the main claims made in the abstract and introduction accurately reflect the paper's contributions and scope?
    \item[] Answer: \answerYes{}
    \item[] Justification: The abstract and introduction accurately reflect the paper's contributions by proposing a novel parameter sharing technique in MARL (\cref{sec: method}) and claiming its superior performance compared to existing methods, which is supported by the experimental results (\cref{sec: experiments}).
    \item[] Guidelines: 
    \begin{itemize}
        \item The answer NA means that the abstract and introduction do not include the claims made in the paper.
        \item The abstract and/or introduction should clearly state the claims made, including the contributions made in the paper and important assumptions and limitations. A No or NA answer to this question will not be perceived well by the reviewers. 
        \item The claims made should match theoretical and experimental results, and reflect how much the results can be expected to generalize to other settings. 
        \item It is fine to include aspirational goals as motivation as long as it is clear that these goals are not attained by the paper. 
    \end{itemize}

\item {\bf Limitations}
    \item[] Question: Does the paper discuss the limitations of the work performed by the authors?
    \item[] Answer: \answerYes{}
    \item[] Justification: Please see \cref{supp: limitations}.
    \item[] Guidelines:
    \begin{itemize}
        \item The answer NA means that the paper has no limitation while the answer No means that the paper has limitations, but those are not discussed in the paper. 
        \item The authors are encouraged to create a separate "Limitations" section in their paper.
        \item The paper should point out any strong assumptions and how robust the results are to violations of these assumptions (e.g., independence assumptions, noiseless settings, model well-specification, asymptotic approximations only holding locally). The authors should reflect on how these assumptions might be violated in practice and what the implications would be.
        \item The authors should reflect on the scope of the claims made, e.g., if the approach was only tested on a few datasets or with a few runs. In general, empirical results often depend on implicit assumptions, which should be articulated.
        \item The authors should reflect on the factors that influence the performance of the approach. For example, a facial recognition algorithm may perform poorly when image resolution is low or images are taken in low lighting. Or a speech-to-text system might not be used reliably to provide closed captions for online lectures because it fails to handle technical jargon.
        \item The authors should discuss the computational efficiency of the proposed algorithms and how they scale with dataset size.
        \item If applicable, the authors should discuss possible limitations of their approach to address problems of privacy and fairness.
        \item While the authors might fear that complete honesty about limitations might be used by reviewers as grounds for rejection, a worse outcome might be that reviewers discover limitations that aren't acknowledged in the paper. The authors should use their best judgment and recognize that individual actions in favor of transparency play an important role in developing norms that preserve the integrity of the community. Reviewers will be specifically instructed to not penalize honesty concerning limitations.
    \end{itemize}

\item {\bf Theory Assumptions and Proofs}
    \item[] Question: For each theoretical result, does the paper provide the full set of assumptions and a complete (and correct) proof?
    \item[] Answer: \answerNA{} 
    \item[] Justification: This paper does not include theoretical results.
    \item[] Guidelines:
    \begin{itemize}
        \item The answer NA means that the paper does not include theoretical results. 
        \item All the theorems, formulas, and proofs in the paper should be numbered and cross-referenced.
        \item All assumptions should be clearly stated or referenced in the statement of any theorems.
        \item The proofs can either appear in the main paper or the supplemental material, but if they appear in the supplemental material, the authors are encouraged to provide a short proof sketch to provide intuition. 
        \item Inversely, any informal proof provided in the core of the paper should be complemented by formal proofs provided in appendix or supplemental material.
        \item Theorems and Lemmas that the proof relies upon should be properly referenced. 
    \end{itemize}

    \item {\bf Experimental Result Reproducibility}
    \item[] Question: Does the paper fully disclose all the information needed to reproduce the main experimental results of the paper to the extent that it affects the main claims and/or conclusions of the paper (regardless of whether the code and data are provided or not)?
    \item[] Answer: \answerYes{}{} 
    \item[] Justification: The method and implementation details are provided in \cref{sec: method} and \cref{supp: implementation_details}, and the experiment settings and environment details are described in \cref{sec: experiments} and \cref{supp: env_details}. 
    \item[] Guidelines:
    \begin{itemize}
        \item The answer NA means that the paper does not include experiments.
        \item If the paper includes experiments, a No answer to this question will not be perceived well by the reviewers: Making the paper reproducible is important, regardless of whether the code and data are provided or not.
        \item If the contribution is a dataset and/or model, the authors should describe the steps taken to make their results reproducible or verifiable. 
        \item Depending on the contribution, reproducibility can be accomplished in various ways. For example, if the contribution is a novel architecture, describing the architecture fully might suffice, or if the contribution is a specific model and empirical evaluation, it may be necessary to either make it possible for others to replicate the model with the same dataset, or provide access to the model. In general. releasing code and data is often one good way to accomplish this, but reproducibility can also be provided via detailed instructions for how to replicate the results, access to a hosted model (e.g., in the case of a large language model), releasing of a model checkpoint, or other means that are appropriate to the research performed.
        \item While NeurIPS does not require releasing code, the conference does require all submissions to provide some reasonable avenue for reproducibility, which may depend on the nature of the contribution. For example
        \begin{enumerate}
            \item If the contribution is primarily a new algorithm, the paper should make it clear how to reproduce that algorithm.
            \item If the contribution is primarily a new model architecture, the paper should describe the architecture clearly and fully.
            \item If the contribution is a new model (e.g., a large language model), then there should either be a way to access this model for reproducing the results or a way to reproduce the model (e.g., with an open-source dataset or instructions for how to construct the dataset).
            \item We recognize that reproducibility may be tricky in some cases, in which case authors are welcome to describe the particular way they provide for reproducibility. In the case of closed-source models, it may be that access to the model is limited in some way (e.g., to registered users), but it should be possible for other researchers to have some path to reproducing or verifying the results.
        \end{enumerate}
    \end{itemize}

\item {\bf Open access to data and code}
    \item[] Question: Does the paper provide open access to the data and code, with sufficient instructions to faithfully reproduce the main experimental results, as described in supplemental material?
    \item[] Answer: \answerYes{}
    \item[] Justification: The code is available at \url{https://github.com/LXXXXR/Kaleidoscope}.
    \item[] Guidelines:
    \begin{itemize}
        \item The answer NA means that paper does not include experiments requiring code.
        \item Please see the NeurIPS code and data submission guidelines (\url{https://nips.cc/public/guides/CodeSubmissionPolicy}) for more details.
        \item While we encourage the release of code and data, we understand that this might not be possible, so “No” is an acceptable answer. Papers cannot be rejected simply for not including code, unless this is central to the contribution (e.g., for a new open-source benchmark).
        \item The instructions should contain the exact command and environment needed to run to reproduce the results. See the NeurIPS code and data submission guidelines (\url{https://nips.cc/public/guides/CodeSubmissionPolicy}) for more details.
        \item The authors should provide instructions on data access and preparation, including how to access the raw data, preprocessed data, intermediate data, and generated data, etc.
        \item The authors should provide scripts to reproduce all experimental results for the new proposed method and baselines. If only a subset of experiments are reproducible, they should state which ones are omitted from the script and why.
        \item At submission time, to preserve anonymity, the authors should release anonymized versions (if applicable).
        \item Providing as much information as possible in supplemental material (appended to the paper) is recommended, but including URLs to data and code is permitted.
    \end{itemize}

\item {\bf Experimental Setting/Details}
    \item[] Question: Does the paper specify all the training and test details (e.g., data splits, hyperparameters, how they were chosen, type of optimizer, etc.) necessary to understand the results?
    \item[] Answer: \answerYes{}{} 
    \item[] Justification: The experiment settings are provided in \cref{sec: experiments} with details elaborated in \cref{supp: implementation_details,supp: env_details}.
    \item[] Guidelines:
    \begin{itemize}
        \item The answer NA means that the paper does not include experiments.
        \item The experimental setting should be presented in the core of the paper to a level of detail that is necessary to appreciate the results and make sense of them.
        \item The full details can be provided either with the code, in appendix, or as supplemental material.
    \end{itemize}

\item {\bf Experiment Statistical Significance}
    \item[] Question: Does the paper report error bars suitably and correctly defined or other appropriate information about the statistical significance of the experiments?
    \item[] Answer: \answerYes{}{} 
    \item[] Justification: The main results are reported with 95\% confidence error bars in \cref{fig:results_mpe_mamujoco,fig:results_smacv2,fig: ablation}.
    \item[] Guidelines:
    \begin{itemize}
        \item The answer NA means that the paper does not include experiments.
        \item The authors should answer "Yes" if the results are accompanied by error bars, confidence intervals, or statistical significance tests, at least for the experiments that support the main claims of the paper.
        \item The factors of variability that the error bars are capturing should be clearly stated (for example, train/test split, initialization, random drawing of some parameter, or overall run with given experimental conditions).
        \item The method for calculating the error bars should be explained (closed form formula, call to a library function, bootstrap, etc.)
        \item The assumptions made should be given (e.g., Normally distributed errors).
        \item It should be clear whether the error bar is the standard deviation or the standard error of the mean.
        \item It is OK to report 1-sigma error bars, but one should state it. The authors should preferably report a 2-sigma error bar than state that they have a 96\% CI, if the hypothesis of Normality of errors is not verified.
        \item For asymmetric distributions, the authors should be careful not to show in tables or figures symmetric error bars that would yield results that are out of range (e.g. negative error rates).
        \item If error bars are reported in tables or plots, The authors should explain in the text how they were calculated and reference the corresponding figures or tables in the text.
    \end{itemize}

\item {\bf Experiments Compute Resources}
    \item[] Question: For each experiment, does the paper provide sufficient information on the computer resources (type of compute workers, memory, time of execution) needed to reproduce the experiments?
    \item[] Answer: \answerYes{} 
    \item[] Justification: Please see \cref{supp: infrast}.
    \item[] Guidelines:
    \begin{itemize}
        \item The answer NA means that the paper does not include experiments.
        \item The paper should indicate the type of compute workers CPU or GPU, internal cluster, or cloud provider, including relevant memory and storage.
        \item The paper should provide the amount of compute required for each of the individual experimental runs as well as estimate the total compute. 
        \item The paper should disclose whether the full research project required more compute than the experiments reported in the paper (e.g., preliminary or failed experiments that didn't make it into the paper). 
    \end{itemize}
    
\item {\bf Code Of Ethics}
    \item[] Question: Does the research conducted in the paper conform, in every respect, with the NeurIPS Code of Ethics \url{https://neurips.cc/public/EthicsGuidelines}?
    \item[] Answer: \answerYes{}{} 
    \item[] Justification: \answerNA{}{}
    \item[] Guidelines:
    \begin{itemize}
        \item The answer NA means that the authors have not reviewed the NeurIPS Code of Ethics.
        \item If the authors answer No, they should explain the special circumstances that require a deviation from the Code of Ethics.
        \item The authors should make sure to preserve anonymity (e.g., if there is a special consideration due to laws or regulations in their jurisdiction).
    \end{itemize}

\item {\bf Broader Impacts}
    \item[] Question: Does the paper discuss both potential positive societal impacts and negative societal impacts of the work performed?
    \item[] Answer: \answerNA{}{} 
    \item[] Justification: This paper presents work whose goal is to advance the field of Reinforcement Learning. There are many potential societal consequences of our work, none which we feel must be specifically highlighted here.
    \item[] Guidelines:
    \begin{itemize}
        \item The answer NA means that there is no societal impact of the work performed.
        \item If the authors answer NA or No, they should explain why their work has no societal impact or why the paper does not address societal impact.
        \item Examples of negative societal impacts include potential malicious or unintended uses (e.g., disinformation, generating fake profiles, surveillance), fairness considerations (e.g., deployment of technologies that could make decisions that unfairly impact specific groups), privacy considerations, and security considerations.
        \item The conference expects that many papers will be foundational research and not tied to particular applications, let alone deployments. However, if there is a direct path to any negative applications, the authors should point it out. For example, it is legitimate to point out that an improvement in the quality of generative models could be used to generate deepfakes for disinformation. On the other hand, it is not needed to point out that a generic algorithm for optimizing neural networks could enable people to train models that generate Deepfakes faster.
        \item The authors should consider possible harms that could arise when the technology is being used as intended and functioning correctly, harms that could arise when the technology is being used as intended but gives incorrect results, and harms following from (intentional or unintentional) misuse of the technology.
        \item If there are negative societal impacts, the authors could also discuss possible mitigation strategies (e.g., gated release of models, providing defenses in addition to attacks, mechanisms for monitoring misuse, mechanisms to monitor how a system learns from feedback over time, improving the efficiency and accessibility of ML).
    \end{itemize}
    
\item {\bf Safeguards}
    \item[] Question: Does the paper describe safeguards that have been put in place for responsible release of data or models that have a high risk for misuse (e.g., pretrained language models, image generators, or scraped datasets)?
    \item[] Answer: \answerNA{}{} 
    \item[] Justification: The experiments in this work are conducted in simulated game environments, thereby presenting minimal risk of misuse.
    \item[] Guidelines:
    \begin{itemize}
        \item The answer NA means that the paper poses no such risks.
        \item Released models that have a high risk for misuse or dual-use should be released with necessary safeguards to allow for controlled use of the model, for example by requiring that users adhere to usage guidelines or restrictions to access the model or implementing safety filters. 
        \item Datasets that have been scraped from the Internet could pose safety risks. The authors should describe how they avoided releasing unsafe images.
        \item We recognize that providing effective safeguards is challenging, and many papers do not require this, but we encourage authors to take this into account and make a best faith effort.
    \end{itemize}

\item {\bf Licenses for existing assets}
    \item[] Question: Are the creators or original owners of assets (e.g., code, data, models), used in the paper, properly credited and are the license and terms of use explicitly mentioned and properly respected?
    \item[] Answer: \answerYes{}{} 
    \item[] Justification: The papers corresponding to the environments used are cited in \cref{sec: experiments}, and the codebases utilized are listed in \cref{suppl： suppl_network_arch_hyper,supp: env_details}.
    \item[] Guidelines:
    \begin{itemize}
        \item The answer NA means that the paper does not use existing assets.
        \item The authors should cite the original paper that produced the code package or dataset.
        \item The authors should state which version of the asset is used and, if possible, include a URL.
        \item The name of the license (e.g., CC-BY 4.0) should be included for each asset.
        \item For scraped data from a particular source (e.g., website), the copyright and terms of service of that source should be provided.
        \item If assets are released, the license, copyright information, and terms of use in the package should be provided. For popular datasets, \url{paperswithcode.com/datasets} has curated licenses for some datasets. Their licensing guide can help determine the license of a dataset.
        \item For existing datasets that are re-packaged, both the original license and the license of the derived asset (if it has changed) should be provided.
        \item If this information is not available online, the authors are encouraged to reach out to the asset's creators.
    \end{itemize}

\item {\bf New Assets}
    \item[] Question: Are new assets introduced in the paper well documented and is the documentation provided alongside the assets?
    \item[] Answer: \answerNA{} 
    \item[] Justification: \answerNA{}{}
    \item[] Guidelines:
    \begin{itemize}
        \item The answer NA means that the paper does not release new assets.
        \item Researchers should communicate the details of the dataset/code/model as part of their submissions via structured templates. This includes details about training, license, limitations, etc. 
        \item The paper should discuss whether and how consent was obtained from people whose asset is used.
        \item At submission time, remember to anonymize your assets (if applicable). You can either create an anonymized URL or include an anonymized zip file.
    \end{itemize}

\item {\bf Crowdsourcing and Research with Human Subjects}
    \item[] Question: For crowdsourcing experiments and research with human subjects, does the paper include the full text of instructions given to participants and screenshots, if applicable, as well as details about compensation (if any)? 
    \item[] Answer: \answerNA{} 
    \item[] Justification: \answerNA{}
    \item[] Guidelines:
    \begin{itemize}
        \item The answer NA means that the paper does not involve crowdsourcing nor research with human subjects.
        \item Including this information in the supplemental material is fine, but if the main contribution of the paper involves human subjects, then as much detail as possible should be included in the main paper. 
        \item According to the NeurIPS Code of Ethics, workers involved in data collection, curation, or other labor should be paid at least the minimum wage in the country of the data collector. 
    \end{itemize}

\item {\bf Institutional Review Board (IRB) Approvals or Equivalent for Research with Human Subjects}
    \item[] Question: Does the paper describe potential risks incurred by study participants, whether such risks were disclosed to the subjects, and whether Institutional Review Board (IRB) approvals (or an equivalent approval/review based on the requirements of your country or institution) were obtained?
    \item[] Answer: \answerNA{} 
    \item[] Justification: \answerNA{}
    \item[] Guidelines:
    \begin{itemize}
        \item The answer NA means that the paper does not involve crowdsourcing nor research with human subjects.
        \item Depending on the country in which research is conducted, IRB approval (or equivalent) may be required for any human subjects research. If you obtained IRB approval, you should clearly state this in the paper. 
        \item We recognize that the procedures for this may vary significantly between institutions and locations, and we expect authors to adhere to the NeurIPS Code of Ethics and the guidelines for their institution. 
        \item For initial submissions, do not include any information that would break anonymity (if applicable), such as the institution conducting the review.
    \end{itemize}

\end{enumerate}

\end{document}